\documentclass[jimaging,article,accept,pdftex,moreauthors]{Definitions/mdpi} 
\firstpage{1} 
\pubvolume{12}
\issuenum{1}
\articlenumber{30}
\pubyear{2026}
\copyrightyear{2026}
\externaleditor{Constantine Kotropoulos} 
\datereceived{4 December 2025} 
\daterevised{22 December 2025} 
\dateaccepted{4 January 2026} 
\datepublished{7 January 2026 } 
\usepackage{xurl}
\def\UrlBreaks{%
\do\/%
\do\a\do\b\do\c\do\d\do\e\do\f\do\g\do\h\do\i\do\j\do\k\do\l%
\do\m\do\n\do\o\do\p\do\q\do\r\do\s\do\t\do\u\do\v\do\w\do\x\do\y\do\z%
\do\A\do\B\do\C\do\D\do\E\do\F\do\G\do\H\do\I\do\J\do\K\do\L%
\do\M\do\N\do\O\do\P\do\Q\do\R\do\S\do\T\do\U\do\V\do\W\do\X\do\Y\do\Z%
\do0\do1\do2\do3\do4\do5\do6\do7\do8\do9\do=\do/\do.\do:%
\do\*\do\-\do\~\do\'\do\"\do\-}

\Title{From Visual to Multimodal: Systematic Ablation of Encoders and Fusion Strategies in Animal Identification}

\Author{{Vasiliy Kudryavtsev $^{1}$\orcidA{},} 
 Kirill Borodin $^{1,}$*\orcidB{}, German Berezin $^{1}$\orcidC{}, Kirill Bubenchikov $^{2}$\orcidD{}, Grach Mkrtchian $^{1,}$*\orcidE{}\linebreak and Alexander Ryzhkov $^{2}$\orcidF{}}

\AuthorNames{Vasiliy Kudryavtsev, Kirill Borodin, German Berezin, Kirill Bubenchikov, Grach Mkrtchian and Alexander Ryzhkov}

\address{%
$^{1}$ \quad Faculty of IT, Technical {University} 
 of Communication and {Informatics, Moscow, Russia, 111024;} 
 {v.d.kudryavcev@mtuci.ru (V.K.); g.a.berezin@edu.mtuci.ru (G.B.)} 
\\
$^{2}$ \quad AI lab, {Avito, Moscow, Russia, 125196;} 
 {kirill8697@gmail.com (K.B.); alexmryzhkov@gmail.com (A.R.)}\\
}

\corres{Correspondence: k.n.borodin@mtuci.ru (K.B.); g.m.mkrtchyan@mtuci.ru (G.M.)}

\abstract{Automated animal identification is a practical task for reuniting lost pets with their owners, yet current systems often struggle due to limited dataset scale and reliance on unimodal visual cues. This study introduces a multimodal verification framework that enhances visual features with semantic identity priors derived from synthetic textual descriptions. We constructed a massive training corpus of 1.9 million photographs covering 695,091~unique animals to support this investigation. Through systematic ablation studies, we identified SigLIP2-Giant and E5-Small-v2 as the optimal vision and text backbones. We further evaluated fusion strategies ranging from simple concatenation to adaptive gating to determine the best method for integrating these modalities. Our proposed approach utilizes a gated fusion mechanism and achieved a Top-1 accuracy of 84.28\% and an Equal Error Rate of 0.0422 on a comprehensive test protocol. These results represent an 11\% improvement over leading unimodal baselines and demonstrate that integrating synthesized semantic descriptions significantly refines decision boundaries in large-scale pet re-identification.\\\\ \textbf{Conference update.} This work was accepted to the FGVC13 Workshop at CVPR 2026. We thank Vadim Ezhov and Nikolai Shcheglov for their help with the poster presentation and demonstration at the conference.
}


\begin{document}


\section{Introduction}
\subsection{{Context and Relevance}}

Individual animal identification is a notable problem across multiple domains that demand reliable and automated solutions~\citep{study_jis, Vidal2021Perspectives}. Reunification of lost pets~\citep{Weiss2012LostPets}, wildlife conservation tracking~\citep{Vidal2021Perspectives, Zheng2022Wild}, livestock management~\citep{Zheng2022Wild, 10.5187/jast.2025.e4}, and veterinary care~\citep{Vidal2021Perspectives, study_jis} all depend on accurate identification systems. Traditional approaches have developed physical markers such as microchips and collar tags~\citep{Lord2010, McGreevy2019} to address this need, yet these methods introduce meaningful practical limitations. Physical markers can fall out, break, or become lost during the animal's lifetime~\citep{Lord2010, Lancaster2015microchip}, rendering them unstable in real-world scenarios. These constraints have motivated exploration of alternative identification methods that do not depend on auxiliary physical devices~\citep{https://doi.org/10.1111/2041-210X.14278, Vidal2021Perspectives,Zheng2022Wild}.

Visual recognition systems offer a practical alternative that circumvents the limitations of physical markers by leveraging the unique biometric features inherent to each \mbox{animal \citep{https://doi.org/10.1111/faf.12861, Petso2022Review}.} Rather than relying on attached or implanted devices, visual-based systems capture facial and body characteristics that remain stable throughout an animal's life~\citep{https://doi.org/10.1111/jzo.13009, Meenakshisundaram2021WhaleSharks}. This approach proves particularly valuable in field scenarios where physical markers may be inaccessible or have been removed~\citep{https://doi.org/10.1111/faf.12861, McCutcheon2025_TrendsMarkRecapture}. Deep learning techniques have enabled scalable processing of visual information by allowing systems to extract and compare biometric features from camera images without human intervention~\citep{revisitinganimalphotoidentification2021, doi:10.1126/sciadv.aaw0736}.

Deep learning-based visual identification systems provide noticeable advantages for animal identification tasks~\citep{study_jis, Vidal2021Perspectives}. These approaches are non-invasive, requiring only image capture rather than physical contact or device implantation~\citep{https://doi.org/10.1111/2041-210X.14278, mougeot2019deeplearningapproachdogface,Ferreira2020}. The methods are inherently scalable, as they can process multiple subjects simultaneously and adapt to new individuals without requiring system reconfiguration~\citep{Romero_Ferrero2019, sakamoto2023markerless, revisitinganimalphotoidentification2021}. In contrast to breed-level classification systems, which categorize animals into predefined species or breed categories, individual identification systems must distinguish between unique subjects within a population~\citep{Vidal2021Perspectives, study_jis, hou2025openanimals}. This more granular level of recognition enables finer-grained management and tracking across diverse real-world deployment scenarios~\citep{optimizing, Zheng2022Wild}.

However, the development of robust visual identification systems faces considerable technical challenges~\citep{hou2025openanimals, optimizing}. Systems must contend with limited training datasets, where the number of individual animals with sufficient labeled image samples remains constrained~\citep{optimizing, Nepovinnykh2022SealID}. Annotation quality and consistency present additional difficulties, particularly when labels are collected across multiple sources or by different annotators~\citep{Nepovinnykh2022SealID, Cermak2024WildlifeDatasets}. A primary limitation emerges from model generalization failures on new animals not encountered during training~\citep{hou2025openanimals, adam2025wildlifereid}. Models trained on specific populations may fail to identify individuals from different environments, with varying lighting conditions, camera angles, or image quality~\citep{ODO202589, terra_incognito}. These generalization failures compromise system reliability in production environments where the model must operate across diverse conditions without retraining~\citep{hou2025openanimals, 10943784}.
Research addressing individual animal identification therefore requires systematic investigation of how to build systems that maintain robust performance across these diverse real-world conditions~\citep{study_jis, hou2025openanimals}. This investigation must encompass dataset construction strategies that capture natural variation~\citep{Cermak2024WildlifeDatasets, otarashvili2024multispeciesanimalreidusing}, architectural choices that generalize effectively~\citep{hou2025openanimals, 9096925, https://doi.org/10.1111/2041-210X.14278}, and evaluation protocols that measure performance realistically~\citep{hou2025openanimals, optimizing}. Understanding which approaches enable reliable identification across different populations, environments, and deployment scenarios~\citep{optimizing} remains essential for enabling widespread adoption of visual identification systems in practical applications~\citep{hou2025openanimals, study_jis}.

\subsection{{Research Problem and Associated Challenges}}

Individual animal identification differs fundamentally from breed-level classification and species recognition tasks~\citep{study_jis}. While classification systems assign animals to predefined categories based on taxonomic or morphological characteristics~\citep{Ratnasingham2013BIN}, individual identification requires distinguishing between unique subjects within a population~\citep{Vidal2021Perspectives, Pereira2022Individual}. This distinction is not merely a matter of granularity; it represents a qualitatively different problem that demands more discriminative feature learning and more demanding evaluation protocols \citep{Vidal2021Perspectives}. Individual identification systems must capture and leverage subtle inter-individual variations~\citep{Paudel2025Advancements} that breed classifiers explicitly ignore~\citep{Vidal2021Perspectives}, making the task substantially more challenging and consequently more valuable for practical applications~\citep{HOU2020108414}.

The challenge of individual identification arises partly from the inherent variability in animal appearance across different observations~\citep{optimizing}. The same animal exhibits considerable visual variation depending on acquisition conditions, body posture, age, and physical state. Lighting conditions significantly affect feature visibility and contrast, while camera viewpoint influences which body regions are visible and how features appear in projection~\citep{terra_incognito}. Age-related changes in appearance, such as fur coloration or facial feature development, introduce temporal variation that models must accommodate~\citep{McCutcheon2025_TrendsMarkRecapture}. These factors combine to create substantial intra-individual variation that can rival or exceed inter-individual differences, complicating the learning of discriminative features.

The availability and quality of training data present notable constraints on system development~\citep{shinoda2024petfacelargescaledatasetbenchmark}. Existing animal identification datasets remain limited in scale compared to human face recognition benchmarks, restricting the diversity of individuals and conditions represented in training corpora~\citep{study_jis}. Beyond scale, annotation quality and consistency become relevant concerns in metric learning frameworks, where individual-level labels directly affect learning outcomes~\citep{ferreira2023pseudolabelingsemisupervisedlearning}. A single mislabeled image may corrupt multiple training triplets during learning, propagating errors throughout the optimization \mbox{process~\citep{Hoffer2015TripletNetwork}.} This sensitivity to label quality means that datasets collected from heterogeneous sources or with varying annotation standards may contain systematic noise that degrades model performance, even if the total number of labeled samples appears sufficient~\citep{Song2025LabelNoiseSurvey}.

Framing animal identity verification as a unified text-image classification problem leverages complementary information: the image contributes fine visual detail, while the text contributes identity-conditioned priors, yielding more separable decision boundaries than image-only models under pose, illumination, and viewpoint variation~\citep{radford2021learning,azizi2023deeplearningpetidentificationusing}. A shared representation space that jointly embeds images and identity text~\citep{zhou2025learningitemrepresentationsdirectly} lets the model highlight stable, discriminative markers linked to the described individual~\citep{huang2025vicp}, while attenuating nuisance variation unrelated to identity~\citep{huang2025vicp}. Under a closed-set protocol where the set of candidate individuals is predefined, this unified space supports a straightforward classifier module and calibrated thresholds for identity recognition, enabling clean comparisons to image-only baselines and making the effect of multimodal fusion easy to quantify~\citep{10943784}.

Optimization should align image and text for the same individual and push apart representations of different individuals using contrastive or metric learning objectives over the shared space~\citep{zhai2023sigmoid}, while preserving strong unimodal encoders so performance remains reliable if one modality is weak. Training can progress from unimodal warm starts to joint alignment, with balanced sampling across individuals to maintain uniform learning pressure and regularization that encourages tight within-identity clusters across changing capture conditions. For assessment, closed-set evaluations allow rigorous measurement of accuracy, calibration, and confusion patterns; ablations make clear when the unified text-image representation materially improves identity discrimination relative to unimodal baselines~\citep{schroff2015facenet}.

\subsection{{Field Snapshot}}

Early animal identification research operated under severe data constraints that fundamentally limited research directions and methodological choices~\citep{study_jis}. Datasets typically contained fewer than 100 individual animals, a scale insufficient for training robust identification models~\citep{Petso2022Review}. These constraints forced the research community toward breed-level classification and species recognition as more tractable alternatives, leaving individual identification largely unexplored despite its greater practical value and technical \mbox{complexity~\citep{Vidal2021Perspectives}.}

Foundational datasets such as \texttt{Stanford Dogs}~\citep{khosla2011noveldatasetfinegrainedimagecategorization} and \texttt{Oxford-IIIT Pet}~\citep{parkhi2012catsdogs} established important benchmarks for fine-grained animal classification and drove methodological advances in deep learning for animal visual analysis. However, these datasets were explicitly designed for breed recognition rather than individual identification. Their limited scope created a persistent research gap between breed classification and individual identification that persisted for nearly a decade, during which the two problems developed as largely separate research tracks with different datasets, evaluation protocols, and technical approaches~\citep{Vidal2021Perspectives}.

The emergence of individual-identification datasets marked a paradigm shift in research objectives~\citep{Vidal2021Perspectives}. \texttt{DogFaceNet}~\citep{mougeot2019deeplearningapproachdogface} pioneered this direction by demonstrating the feasibility of adapting face-recognition techniques to animal domains, establishing the viability of individual animal identification using facial features. Although this pioneer work operated at a limited scale, its primary contribution lay in opening a new research trajectory that motivated subsequent work toward larger and more comprehensive individual-\mbox{identification systems.}

Multi-species approaches like \texttt{AnimalWeb}~\citep{khan2020animalweblargescalehierarchicaldataset} represent attempts to improve model generalization across taxonomic boundaries by learning shared feature representations. These approaches face inherent challenges in balancing species diversity against adequate per-species representation. \texttt{AnimalWeb}'s hierarchical taxonomy mirrors biological classifications but does not necessarily align with optimal feature hierarchies for computer vision tasks, creating a mismatch between biological organization and computational \mbox{optimization objectives.}

Different biometric modalities offer distinct characteristics that shape their practical applicability~\citep{tharwat2018animalbiometricssurvey}. Nose-print patterns remain stable throughout an animal's lifetime and exhibit unique ridge structures~\citep{choi2021studyviabilitycaninesepattern}, yet their utility is constrained by acquisition requirements that demand close-range imaging and subject cooperation~\citep{chan2024dognoseprintrecognitionbased}. Facial features enable recognition from typical camera distances and work across diverse viewing conditions~\citep{Ferreira2020}, though they exhibit greater variation across time and conditions compared to specialized biometric patterns. Body markings provide globally unique signals but exhibit variability due to seasonal changes, injuries, and growth~\citep{ODO202589}, requiring models to handle substantial intra-individual variation.

Animal identification research has adapted techniques from human face recognition, particularly loss functions and metric learning frameworks that optimize for discriminative embedding spaces~\citep{schroff2015facenet}. However, this adaptation introduces significant challenges because animal faces present distinct characteristics from human faces, including greater intra-individual variation, different facial structural constraints, and species-specific features that may not align with architectures developed for human recognition~\citep{clapham2020automatedfacialrecognitionwildlife}.

Recent developments have expanded the landscape of available methods and datasets. Large-scale initiatives such as \texttt{WildlifeReID-10k}~\citep{adam2025wildlifereid} and community-curated multispecies collections~\citep{otarashvili2024multispeciesanimalreidusing} now enable investigation of diverse animal species under real-world conditions. Semi-supervised learning approaches address limited labeled data through pseudo-labeling and consistency regularization~\citep{ferreira2023pseudolabelingsemisupervisedlearning}, while cross-species transfer learning attempts to leverage features learned on one species for identification in others~\citep{biggs2020deepcrossspeciesfeaturelearning}. The introduction of \texttt{PetFace}~\citep{shinoda2024petfacelargescaledatasetbenchmark}, providing 257,484 unique individuals across multiple species, enables investigation of previously intractable problems such as unseen-individual verification. Open-source tools such as \texttt{WildlifeDatasets}~\citep{Cermak2024WildlifeDatasets} have facilitated standardized evaluation and reproducibility. Vision--language models have recently been adapted for animal-centric tasks~\citep{10943784}, and advances in robust visual encoders such as DINOv2~\citep{oquab2024dinov2learningrobustvisual} and multilingual SigLIP~\citep{tschannen2025siglip2multilingualvisionlanguage} provide powerful foundation models for animal identification. However, larger datasets frequently involve trade-offs with annotation quality and quality control~\citep{Petso2022Review}, introducing systematic noise that may degrade performance in metric-learning frameworks where label integrity directly affects training dynamics~\citep{Hoffer2015TripletNetwork}.

Fundamental questions remain open regarding the optimal balance between dataset scale and annotation quality, the extent to which biometric features transfer across species boundaries~\citep{transferlearning2024animalspecies}, and the specific conditions under which models trained on controlled datasets generalize to real-world deployment scenarios~\citep{terra_incognito}. These questions have not received systematic investigation across standardized evaluation protocols, leaving important gaps in our understanding of how to build reliable identification systems for practical applications~\citep{optimizing}.

\subsection{{Gap and Rationale}}

Different studies on animal identification use different datasets, evaluation metrics, and training protocols, making it difficult to compare results across the literature~\citep{study_jis}. When DogFaceNet~\citep{mougeot2019deeplearningapproachdogface} demonstrates strong performance on one dataset, it is unclear whether this reflects a superior method or simply favorable experimental conditions~\citep{study_jis}. Cross-species transfer learning and breed-specific fine-tuning report conflicting conclusions about which approach works best, and these different methods operate under incomparable evaluation standards. This inconsistency prevents researchers from determining which architectural choices actually improve identification performance~\citep{study_jis}.

Individual animal identification has remained unimodal, relying solely on visual features~\citep{Vidal2021Perspectives}, while human experts naturally use both visual observation and textual descriptions when identifying animals. Veterinarians and shelter workers describe distinctive marks, physical characteristics, and other identifying features alongside visual \mbox{assessment~\citep{Vidal2021Perspectives}.} Vision--language models such as CLIP~\citep{radford2021learning} have proven effective on human-centric tasks~\citep{radford2021learning}, but their potential for animal identification combined with textual descriptions remains unexplored~\citep{10.1145/3746027.3758249}. This represents an opportunity to leverage a naturally multimodal signal that existing systems do not exploit~\citep{10.1145/3746027.3758249, Vidal2021Perspectives}.

Currently, few works systematically examine how different text encoder architectures and fusion strategies affect animal verification performance~\citep{10.1007/978-3-031-92387-6_2}. While vision encoder comparisons exist in the literature, text encoding choices for animal descriptions remain uninvestigated~\citep{10.5187/jast.2025.e4}. Different transformer variants, embedding dimensions, and fusion mechanisms such as adaptive gating and cross-attention have not been evaluated in the animal identification context~\citep{10.1007/978-3-031-92387-6_2}. Without systematic ablation studies isolating each component's contribution, practitioners lack guidance on which combinations to use~\citep{10.5187/jast.2025.e4}. Table~\ref{tab:prior_work_gaps} summarizes representative prior studies, their main limitations, and how the present work addresses these gaps.

\begin{table}[H]
\caption{{Summary} 
 of key gaps in prior work and how the present study addresses them.\label{tab:prior_work_gaps}}
 \begin{adjustwidth}{-\extralength}{0cm}
  \begin{tabularx}{\fulllength}{@{}>{\raggedright\arraybackslash}m{4cm}@{}>{\raggedright\arraybackslash}m{3.8cm}@{}>{\raggedright\arraybackslash}m{10.6cm}}
        \toprule
        \textbf{Gap} & \textbf{Representative Works} & \textbf{How This Work Addresses It} \\
        \midrule
        No unified evaluation~protocol & \citep{mougeot2019deeplearningapproachdogface, study_jis, 10.5187/jast.2025.e4, Vidal2021Perspectives} & We combine five large training datasets into a unified corpus and define a single standardized training and evaluation protocol used for all backbones and configurations. \\
          \midrule
        Unimodality & \citep{mougeot2019deeplearningapproachdogface, 10.1007/978-3-031-92387-6_2, 10.5187/jast.2025.e4, Vidal2021Perspectives} & We introduce a multimodal pipeline that augments visual embeddings with identity-level textual descriptions. \\
          \midrule
        No systematic ablation~studies & \citep{mougeot2019deeplearningapproachdogface, study_jis, 10.1007/978-3-031-92387-6_2, 10.5187/jast.2025.e4, Vidal2021Perspectives} & We perform controlled ablations over multiple vision encoders, text encoders, and training configurations, isolating their contributions to verification performance on a common benchmark suite. \\
          \midrule
        No controlled comparison~of multimodal fusion strategies &~\citep{mougeot2019deeplearningapproachdogface, study_jis, 10.5187/jast.2025.e4, 10.1145/3746027.3758249, Vidal2021Perspectives} & We implement and compare several fusion mechanisms  under identical settings, identifying effective designs for animal verification. \\
        \bottomrule
    \end{tabularx}
  \end{adjustwidth}
\end{table}


To summarize, three gaps limit progress toward practical, comparable animal verification {systems:} 
\begin{enumerate}
\item[(i)] The lack of a unified evaluation protocol that enables fair, controlled comparison of vision encoders across heterogeneous datasets; 
\item[(ii)] The lack of multimodal verification pipelines that exploit identity-level textual cues, despite their natural use by human experts; and 
\item[(iii)] The absence of systematic ablations of text encoders and fusion strategies in the animal identification setting.
\end{enumerate}

Accordingly, what is needed is a controlled experimental framework that standardizes training and evaluation across backbones, and a principled multimodal design that quantifies when and how text improves individual verification. 
This paper fills these gaps by establishing a unified large-scale training corpus and standardized protocol, and by conducting encoder- and fusion-level ablations for image--text animal verification using synthesized identity descriptors.

\subsection{{Contribution}}

Individual animal identification is a critical problem in computer vision with applications in pet reunification, wildlife conservation, and veterinary management~\citep{Vidal2021Perspectives,study_jis}. Traditional physical markers like tags and microchips have limitations in accessibility and durability~\citep{McGreevy2019,Lord2010,Lancaster2015microchip}. Deep learning-based visual identification offers a non-invasive alternative, but existing work faces challenges in dataset scale, annotation quality, and real-world generalization~\citep{shinoda2024petfacelargescaledatasetbenchmark}.

This paper makes the following contributions:

\begin{itemize}
\item \textbf{{Unified} 
large-scale corpus and protocol.} We construct a unified training corpus by combining Pet911.ru and Telegram data with established benchmarks (Dogs-World~\citep{dogs_world_kaggle}, LCW~\citep{lcw_cats}, PetFace~\citep{shinoda2024petfacelargescaledatasetbenchmark}), totaling 695,091 unique animals and \mbox{1,904,157 photographs,} and define a standardized evaluation protocol to enable \mbox{fair comparisons.}
\item \textbf{Controlled vision backbone ablation.} We benchmark multiple vision encoders under identical training conditions (losses, optimization, schedule, and metrics), ensuring that observed differences reflect architectural choices rather than experimental \mbox{variation~\citep{Petso2022Review}.}
\item \textbf{Multimodal verification with systematic ablations.} We introduce an image--text animal verification pipeline that augments visual embeddings with synthesized identity descriptions~\citep{mougeot2019deeplearningapproachdogface,clapham2020automatedfacialrecognitionwildlife}, and we systematically ablate text encoder architectures~\citep{mougeot2019deeplearningapproachdogface,clapham2020automatedfacialrecognitionwildlife} and multimodal fusion strategies (including cross-attention and gating) to provide practical guidance on effective combinations. Different fusion strategies are compared to measure how multimodal integration improves verification performance relative to image-only baselines~\citep{radford2021learning,10943784}.
\end{itemize}

Together, these contributions address the lack of comparability across studies and the absence of systematic multimodal investigations in animal identification, while providing a reproducible foundation for settings where both images and descriptive text are available.

\section{Materials and Methods}

\subsection{{Data}}

\textbf{Pet911 Dataset.} We constructed a dataset through automated web scraping of {the} 
 \url{pet911.ru} platform (accessed on 1 November 2025), a Russian service for lost and found pet announcements. The parsing implementation employs BeautifulSoup for Hypertext Markup Language (HTML) processing and requests library for HyperText Transfer Protocol (HTTP) communication with error handling. The system navigates catalog pages using pagination detection algorithms to identify available content. For each listing, we extracted animal metadata including species classification, descriptive text, and associated photographs. We filtered listings to retain only animals with at least two photographs per individual.  Downloaded images underwent validation for format consistency, with automatic conversion of WebP formats to Joint Photographic Experts Group (JPEG) for standardization~\citep{McGreevy2019}. The Pet911 dataset yielded 65,961~photographs representing 22,050 unique animals. Each animal record includes species classification for cats or dogs, textual description, and between 2 and 8 associated photographs. The dataset captures real-world variability in image quality, lighting conditions, and animal poses representative of lost pet scenarios~\citep{Weiss2012LostPets}.

\textbf{Telegram Channel Dataset.} The Telegram dataset construction utilized the Telethon library to access public animal-related channels through the Telegram Application Programming Interface (API).  The system processes message streams from targeted public channels, using keyword matching to identify animal-related content~\citep{study_jis}. Media processing handles both individual photos and grouped albums. The system automatically detects grouped messages and downloads all associated images while maintaining proper file organization.   The Telegram dataset contributed 131,698 photographs from 73,101 unique animals. This source provides complementary data characteristics, including casual photography styles, varied backgrounds, and diverse animal representations not captured in formal lost pet platforms~\citep{Petso2022Review}.

\textbf{Existing Datasets.} Beyond our constructed datasets, we incorporated established benchmarks that represent diverse data collection methodologies and real-world \mbox{scenarios~\citep{Vidal2021Perspectives}.} The Dogs-World dataset~\citep{dogs_world_kaggle} provides 301,342 photographs from 200,458 unique dogs, capturing variations in controlled and semi-controlled environments. The LCW dataset~\citep{lcw_cats} contributes 381,267 photographs representing 140,732 individual animals, expanding the diversity of acquisition conditions and animal populations~\citep{hou2025openanimals}. PetFace, the largest benchmark in our evaluation, contains 1,001,532 photographs representing 257,349 unique animals across multiple species~\citep{shinoda2024petfacelargescaledatasetbenchmark}. For evaluation purposes, we utilized Cat Individual Images~\citep{timost1234_cat_individuals_2019}, which provides 13,542 photographs of 518 individual cats, and DogFaceNet~\citep{mougeot2019deeplearningapproachdogface}, consisting of 8363 photographs from 2483 unique dogs, both serving as controlled test sets for assessing model generalization across different animal populations. These established datasets have been evaluated in prior work and demonstrate the trade-off between dataset scale and annotation quality that characterizes recent progress in animal \mbox{identification research.}

\textbf{Combined Dataset Composition.} The training corpus combines our constructed datasets with established datasets, leveraging comprehensive scale and diversity across multiple data sources~\citep{otarashvili2024multispeciesanimalreidusing}. As presented in Table~\ref{tab:datasets}, our complete dataset contains \mbox{1,904,157 total} photographs representing 695,091 unique animals across cats and dogs. The combination of constructed and established datasets provides a robust foundation for model training and evaluation across diverse scenarios and animal types~\citep{https://doi.org/10.1111/2041-210X.14278,Nepovinnykh2022SealID}.  {In addition to the total number of identities and photos, we report descriptive statistics of the number of photos per identity: \textit{{min} 
} and \textit{max} denote the minimum and maximum number of images available for a single identity, while \textit{mean}, \textit{med} (median), and \textit{std} denote the average, median, and standard deviation of images per identity, respectively. These statistics quantify the per-identity sample-count distribution and highlight differences in data balance across sources.} Training uses balanced sampling to ensure equal representation of identities within each batch, addressing class imbalance issues where some animals have significantly more photos than others~\citep{Ferreira2020,Zheng2022Wild}.

\textbf{Data Preprocessing.} We evaluated the impact of automated animal detection preprocessing on model performance~\citep{sakamoto2023markerless,Romero_Ferrero2019}. The baseline experiment uses the original dataset without additional preprocessing. A second configuration incorporates YOLO12~\citep{tian2025yolov12} object detection to crop animal regions before feature extraction, testing whether explicit localization improves identification performance~\citep{tian2025yolov12}.

\begin{table}[H]
\caption{Dataset Composition and Statistics.\label{tab:datasets}}
  \begin{adjustwidth}{-\extralength}{0cm}
    \begin{tabular*}{\fulllength}{@{\extracolsep{\fill}} l c c c c c c c c c c }
      \toprule
      \textbf{Dataset} & \textbf{Identities} & \textbf{Photos} & \textbf{Usage} & \textbf{\textit{Min}} & \textbf{\textit{Max}} & \textbf{\textit{Mean}} &   \textbf{\textit{Med}} & \textbf{\textit{Std}}\\
      \midrule
      Pet911 (ours)            & 22,050   & 65,961    & Training & 2  & 10 & 2.99 & 3 & 1.34      \\
      Telegram (ours)         & 73,101   & 131,698   & Training & 2 & 10 & 2.73 & 2 & 1.05      \\
      Dogs-World~\citep{dogs_world_kaggle}        & 200,458  & 301,342   & Training & 2 & 17 & 2.36 & 2 & 0.63     \\
      LCW~\citep{lcw_cats}               & 140,732  & 381,267   & Training & 2 & 6 & 3.71 & 2 & 1.45      \\
      PetFace~\citep{shinoda2024petfacelargescaledatasetbenchmark}           & 257,349  & 1,001,532 & Training & 3 & 11 & 4.58 & 4 & 1.29 \\
      Cat Individual Images~\citep{timost1234_cat_individuals_2019}           & 518      & 13,542    & Evaluation & 6 & 144 & 25.58 & 20 & 10.06       \\
      DogFaceNet~\citep{mougeot2019deeplearningapproachdogface}        & {2483} 
    & 8363     & Evaluation & 2 & 17 & 2.36 & 2 & 0.63       \\
      \bottomrule
    \end{tabular*}
  \end{adjustwidth}
\end{table}

\textbf{Data Composition Ablation Experiments.} We systematically evaluated how different data sources impact model performance through controlled ablation studies. Our primary investigation examined whether incorporating our newly collected Pet911 and Telegram datasets improves identification accuracy compared to training solely on established benchmarks. The PetFace~\cite{shinoda2024petfacelargescaledatasetbenchmark} dataset presents a specific methodological challenge: all images underwent automated face detection, precise alignment, and manual filtering, resulting in a highly controlled distribution that differs substantially from real-world deployment scenarios. To quantify this distribution mismatch effect, we designed three experimental configurations: training without PetFace~\cite{shinoda2024petfacelargescaledatasetbenchmark} to assess performance on unfiltered data, training with the PetFace~\cite{shinoda2024petfacelargescaledatasetbenchmark} training split only following standard protocols, and  training with the complete PetFace dataset to examine whether scale compensates for domain shift.

\textbf{Training and Test Split.} Our experimental framework employs stratified splits maintaining animal identity separation between training and test sets to ensure valid evaluation of model generalization~\citep{terra_incognito}. The training set comprises five different datasets, while the test set contains Cat Individual Images~\citep{timost1234_cat_individuals_2019} and DogFaceNet~\citep{mougeot2019deeplearningapproachdogface}. This configuration prevents data leakage and enables proper assessment of individual identification performance on previously unseen animals~\citep{doi:10.1126/sciadv.aaw0736}.

\subsection{{Vision Encoder Experiments}}\label{vision_encoder_description}

\textbf{Vision Encoder Selection.} We evaluated six pre-trained vision encoders that represent different architectural approaches and pre-training objectives relevant to animal identification tasks. \texttt{CLIP-ViT-Base}~\citep{radford2021learning} combines vision transformer architecture with language-image contrastive learning, enabling models to leverage semantic relationships between visual and textual information. \texttt{SigLIP-Base}~\citep{zhai2023sigmoid} employs sigmoid loss for contrastive learning, offering improved training stability and convergence properties compared to standard softmax-based approaches. \texttt{SigLIP2-Base}~\citep{tschannen2025siglip2multilingualvisionlanguage} represents an updated version of SigLIP with refined training procedures and architectural improvements.  \texttt{SigLIP2-Giant} is a scaled-up variant of the SigLIP2~\citep{tschannen2025siglip2multilingualvisionlanguage}  architecture to giant scale with optimized training and higher input resolution, providing state-of-the-art comparable visual representations through massive model capacity and enhanced fine-grained detail capture. \texttt{DINOv2-Small}~\citep{oquab2024dinov2learningrobustvisual} uses self-supervised learning on diverse image collections without language supervision, enabling the discovery of task-agnostic visual features that generalize across domains. \texttt{Zer0int CLIP-L} provides a large-scale CLIP variant with geometric mean pooling aggregation, offering increased model capacity and refined feature aggregation strategies. These diverse encoders enable systematic investigation of how different pre-training objectives influence feature quality for individual animal identification. Table~\ref{tab:vision_complexity} further summarizes the computational characteristics of each vision backbone, including the number of parameters, the number of multiply–add operations (Mult-Adds), and the peak inference VRAM footprint in megabytes (batch size = 1), the average training time per epoch in seconds, and the inference throughput measured in images per second.

\begin{table}[H]
\caption{Vision Encoders' computational characteristics.\label{tab:vision_complexity}}
    \begin{tabular*}{\textwidth}{@{\extracolsep{\fill}}l c c c c c}
      \toprule
      \textbf{Configuration} & \textbf{Parameters} & \textbf{Mult-Adds} & \textbf{VRAM} & \textbf{Epoch Time} & \textbf{Throughput}  \\
      \midrule
      CLIP-ViT-Base & 151,277,313 & 201,094,656 & 375.24 & 7641 & 211\\
      DINOv2-Small & 22,056,576 & 79,195,392 & 201.66 & 2497 & 224 \\
      SigLIP-Base & 203,155,970 & 205,686,528 & 500.96 & 4354 & 151  \\
      SigLIP2-Base & 375,187,970 & 205,686,528 &  500.96 & 4271 & 160\\
      Zer0int CLIP-L & 427,616,513 & 457,503,744 & 1696.21 & {10,041} 
 & 66 \\
      SigLIP2-Giant & 1,871,885,426 & 1,833,395,712 & 7394.99 & 64,641 & 10 \\
      \bottomrule
    \end{tabular*}
\end{table}

\textbf{Training Configuration.} All vision encoders undergo identical training procedures to ensure fair comparison across different architectural approaches. Training employs a batch size of 116 samples structured as 58 unique animal identities with 2 photographs each, ensuring balanced identity representation within every training iteration. The learning rate is fixed at {1 $\times$ 10$^{-4}$} 
 with Adam optimization using default parameters, providing consistent gradient updates across all encoders. Training proceeds for 10 epochs across all experiments, establishing a standardized training duration that allows sufficient convergence while maintaining consistent computational requirements. This configuration enables assessment of encoder performance under identical learning conditions, revealing which architectural choices and pre-training objectives produce superior feature representations for \mbox{animal identification.}

\textbf{Transfer Learning Strategy.} All vision encoders utilize transfer learning by freezing lower layers while unfreezing only the final five transformer blocks during training. This approach preserves general-purpose visual features learned during large-scale pre-training on diverse image datasets, while enabling adaptation to animal identification tasks through fine-tuning higher-level features. Freezing early layers maintains foundational feature patterns that remain useful across different domains, reducing catastrophic forgetting and improving convergence speed. Unfrozen final blocks allow the model to learn animal-specific feature representations that discriminate between individual subjects. This balance between preservation and adaptation leverages the benefits of pre-trained models while enabling task-specific optimization without requiring extensive training resources.

\textbf{Sampling.} Training employs a balanced identity sampler that ensures equal representation of animal identities within each batch. This sampling strategy guarantees that each of the 58 identities appears exactly twice per batch, regardless of how many total photographs each identity possesses. This approach directly addresses class imbalance issues inherent in animal identification datasets, where some animals have many photographs while others have few. Balanced sampling improves convergence by preventing the model from biasing toward frequently-represented identities and ensures that less-represented animals contribute equally to gradient updates.

\subsection{{Text Generation}}

For text generation, we utilized \texttt{Qwen3-VL}~\citep{bai2025qwen25vltechnicalreport}, a state-of-the-art multimodal large language model developed with capabilities for both vision and language understanding. \texttt{Qwen3-VL} is designed to interpret visual inputs and generate highly structured descriptive text, making it well-suited for automated annotation tasks. Because individual manual textual descriptions are unavailable for each sample in our dataset, the model is employed to produce consistent, standardized descriptions across our entire collection. This approach enables scalable annotation and minimizes potential variability or human bias associated with real text data, ensuring each sample receives uniform descriptive metadata tailored to our verification objectives.

\subsection{{Text Encoder Experiments}}

Ablation studies of text encoder architectures are conducted to ensure methodological consistency and control across modalities. We evaluate \texttt{E5-Base}~\citep{wang2024textembeddingsweaklysupervisedcontrastive}, a transformer-based model tailored for semantic retrieval, as well as \texttt{E5-Small}~\citep{wang2024textembeddingsweaklysupervisedcontrastive} and their respective v2 versions (\texttt{E5-Small-v2}~\citep{wang2024textembeddingsweaklysupervisedcontrastive} and \texttt{E5-Base-v2}~\citep{wang2024textembeddingsweaklysupervisedcontrastive}), which provide improved computational efficiency and representational accuracy via updated training procedures. Additionally, our experiments include \texttt{BERT}~\citep{devlin2019bert}, the standard backbone for general-purpose language modeling. All text encoder experiments are trained under identical configurations that mirror those of the vision encoder experiments (Section~\ref{vision_encoder_description}), including batch size, learning rate, optimization protocol, balanced sampling, and a transfer learning strategy based on partial layer freezing. This unified protocol enables fair cross-modality comparisons and isolates the impact of each text encoder on downstream verification performance. Table~\ref{tab:text_complexity} further summarizes the computational characteristics of each text backbone, including the number of parameters, the number of multiply–add operations (Mult‑Adds), the peak inference VRAM footprint in megabytes (batch size = 1), the average training time per epoch in seconds, and the inference throughput measured in images per second.

\begin{table}[H]
\caption{Text encoders' computational characteristics.\label{tab:text_complexity}}
    \begin{tabular*}{\textwidth}{@{\extracolsep{\fill}}l c c c c c}
      \toprule
      \textbf{Configuration} & \textbf{Parameters} & \textbf{Mult-Adds} & \textbf{VRAM} & \textbf{Epoch Time} & \textbf{Throughput}  \\
      \midrule
      BERT &  109,482,240 & 109,482,240 & 519.65 &  669 & 990\\
      E5-Base & 109,482,240 & 109,482,240 & 519.65 & 688 & 985\\
      E5-Base-v2 &  109,482,240 & 109,482,240 & 519.65 & 670 & 963\\
      E5-Small & 33,360,000 & 33,360,000 & 178.26 & 657 & 986  \\
      E5-Small-v2 & 33,360,000 & 33,360,000 & 178.26 & 661 & 997 \\
      \bottomrule
    \end{tabular*}
\end{table}

\subsection{{Multimodal Experiments}}

In three dual-encoder baselines(\texttt{CLIP-ViT-Base + E5-Base-v2}, \texttt{CLIP-ViT-Base + E5-Small-v2} and \texttt{SigLIP2-Giant + E5-Small-v2}), image and text embeddings are first projected into a shared space and then concatenated to form a joint representation. Specifically, we pair \texttt{CLIP-ViT-Base} with either \texttt{E5-Base-v2} or \texttt{E5-Small-v2}, and \texttt{SigLIP2-Giant} with \texttt{E5-Small-v2}, comparing the impact of different vision and text encoders under the same fusion scheme.  We use the second version of the small text encoder (\texttt{E5-Small-v2}) as it provides a better efficiency–quality trade-off in our setting. Empirically, the small text encoder variants achieve higher retrieval performance than the base counterpart, so subsequent experiments focus on E5-Small-v2 as the default text encoder, while BERT-based baselines are omitted due to clearly inferior results discussed in Section~\ref{discussion:text}.

In the cross-attention variants, \texttt{CLIP-ViT-Base + E5-Small-v2 cross-attention} and \texttt{SigLIP2-Giant + E5-Small-v2 + cross-attention} first produce image patch embeddings and text token embeddings, which are then fused by an attention module where text features query the image features. The attended text representations, enriched with information from the corresponding image features, are pooled into a single multimodal embedding that replaces simple concatenation and is used for downstream retrieval.

In the weighted-text variants, \texttt{CLIP-ViT-Base + E5-Small-v2} and \texttt{SigLIP2-Giant + E5-Small-v2} use the same dual-encoder and concatenation scheme as the baselines, but apply a learnable scalar weight to the text embedding before fusion. Image and text features are projected into a shared space, the text embedding is rescaled by this trainable factor, and then concatenated with the image embedding to form the final multimodal representation used for retrieval.

For the gated fusion variant \texttt{SigLIP2-Giant + E5-Small-v2 + gating}, image and text embeddings are first projected into a shared space with separate linear layers and then concatenated. This concatenated vector is passed through a small MLP with softmax over two outputs, yielding normalized weights for the text and image embeddings, which are combined as a weighted sum to obtain the final multimodal representation used \mbox{for retrieval.}

Table~\ref{tab:multimodal_complexity} further summarizes the computational characteristics of each multimodal configuration, reporting the total number of trainable parameters across both encoders and fusion module, the aggregate number of multiply–add operations (Mult‑Adds), the peak inference VRAM footprint in megabytes, the average training time per epoch in seconds, and the inference throughput measured in samples processed per second.

\begin{table}[H]
\caption{Multimodal setups' computational characteristics.\label{tab:multimodal_complexity}}
\small
  \begin{adjustwidth}{-\extralength}{0cm}
    \begin{tabular*}{\fulllength}{@{\extracolsep{\fill}}l c c c c c}
      \toprule
      \textbf{Configuration} & \textbf{Parameters} & \textbf{Mult-Adds} & \textbf{VRAM} & \textbf{Epoch Time} & \textbf{Throughput}   \\
      \midrule
      CLIP-ViT-Base + E5-Base-v2 & 261,415,937 & 311,233,280 & 897.41 & 2813 & 203\\
      CLIP-ViT-Base + E5-Small-v2 & 185,097,089 & 234,914,432 & 555.27 & 2788 & 204\\ 
      CLIP-ViT-Base + E5-Small-v2 + cross-attention &  187,988,865 & 235,704,960 & 588.3 & 2952 & 195\\ 
      CLIP-ViT-Base + E5-Small-v2 + weighted text &  185,097,090 & 234,914,432 & 555.27 & 2905 & 194\\
      SigLIP2-Giant + E5-Small-v2 & 1,907,803,890  & 1,869,314,176 & 7557.53 & 63,959 & 10\\
      SigLIP2-Giant + E5-Small-v2 + cross-attention & 1,907,805,426 & 1,868,265,088 & 7556.02 & 63,296 & 10\\
      SigLIP2-Giant + E5-Small-v2 + weighted text & 1,906,229,491 & 1,867,739,776 & 7551.51 & 64,209 & 10\\
      SigLIP2-Giant + E5-Small-v2 + gating & 1,906,360,948 & 1,867,871,234 & 7552.01 & 63,922 & 10\\
      \bottomrule
    \end{tabular*}
  \end{adjustwidth}
\end{table}

\subsection{{Comparison Methods}}

To assess the performance of the proposed approach, we compare it against several strong models pre-trained for wildlife re-identification and biological taxonomy. These models are utilized as fixed feature extractors without any additional fine-tuning on the target datasets, ensuring that the comparison focuses on the generalizability of their learned representations. Inference is conducted using the same evaluation protocol as employed for our main method to guarantee a fair assessment. 

The comparison includes \texttt{MiewID-msv3}~\cite{WildMe2023}, a specialized feature extractor trained using contrastive learning on a high-quality dataset covering 64 different wildlife species, ranging from terrestrial mammals to aquatic animals.

Additionally, we evaluate three distinct architectures from the MegaDescriptor \mbox{family~\citep{Cermak2024WildlifeDatasets},} which are designed as foundation models for individual animal re-identification: \texttt{MD-T-CNN-288}, which is based on the EfficientNet-B3 convolutional neural network~\citep{tan2019efficientnet}; \texttt{MD-CLIP-336}, which adapts a large Vision Transformer initially pre-trained with CLIP~\citep{radford2021learning}; and \texttt{MD-L-384}, which leverages a Swin Transformer Large backbone~\citep{9710580}. Finally, we include \texttt{BioCLIP}~\citep{stevens2024bioclip}, a biology-focused vision foundation model based on the CLIP \mbox{ViT \citep{radford2021learning} architecture.}

Table~\ref{tab:comparison_complexity} summarizes the computational characteristics of the pre-trained comparison models, including the number of parameters, the number of multiply–add operations (Mult‑Adds), the peak inference VRAM footprint in megabytes (batch size = 1), and the inference throughput measured in images per second, while omitting training time since these models are used only in frozen, inference‑only mode.

\begin{table}[H]
\caption{Comparison methods' computational characteristics.\label{tab:comparison_complexity}}
    \begin{tabular*}{\textwidth}{@{\extracolsep{\fill}}l c c c c}
      \toprule
      \textbf{Configuration} & \textbf{Parameters} & \textbf{Mult-Adds} & \textbf{VRAM} & \textbf{Throughput} \\
      \midrule
      MiewID-msv3 & 51,109,277 & 24,307,036,098 & 891.97 & 159\\
      MD-T-CNN-288 &  12,233,232 & 1,589,219,176 & 211.22 & 353\\
      MD-CLIP-336 &  303,507,456 & 649,129,984 & 2360.4 & 36\\
      MD-L-384 & 195,198,516 & 445,579,776 & 1960.17 & 46\\
      BioCLIP &  86,192,640 & 172,314,624 & 391.57 & 267\\
      \bottomrule
    \end{tabular*}
\end{table}

\subsection{{Loss Function Design}}
The training objective combines two complementary components that jointly optimize the feature space for individual animal identification. 

\textbf{Triplet Loss.} We employ triplet loss~\citep{Hoffer2015TripletNetwork} with margin $\alpha = 0.45$ to encourage separation between different animal identities by penalizing cases where different animals produce similar embeddings. For a triplet consisting of an anchor image $x_a$, a positive image $x_p$ (same animal), and a negative image $x_n$ (different animal), the triplet loss is defined as
\begin{linenomath}
\begin{equation}\label{eq:triplet}
\mathcal{L}_{\text{triplet}} = \max(0, \|f(x_a) - f(x_p)\|_2^2 - \|f(x_a) - f(x_n)\|_2^2 + \alpha)
\end{equation}
\end{linenomath}
where $f(\cdot)$ represents the neural network, $\|\cdot\|_2$ denotes the Euclidean distance, and $\alpha$ establishes the minimum required distance between features from different animals.

\textbf{Intra-Pair Variance Regularization.} We apply intra-pair variance regularization~\citep{Yu2019TupletMarginLoss} to promote consistency across multiple photographs of the same animal. This loss minimizes the variance of similarity scores within both positive pairs (same identity) and negative pairs (different identities), encouraging tighter clustering and more stable decision boundaries. 

For positive pairs with cosine similarity scores $\{s_p^i\}_{i=1}^{N_p}$ and negative pairs with similarity scores $\{s_n^j\}_{j=1}^{N_n}$, the intra-pair variance loss is computed as

\begin{linenomath}
\begin{equation}\label{eq:pos_loss}
\mathcal{L}_{\text{var}}^{\text{pos}} = \max\left(0, (1 - \varepsilon_{\text{pos}}) \bar{s}_p - s_p^i\right)^2
\end{equation}
\end{linenomath}

\begin{linenomath}
\begin{equation}\label{eq:neg_loss}
\mathcal{L}_{\text{var}}^{\text{neg}} = \max\left(0, s_n^j - (1 + \varepsilon_{\text{neg}}) \bar{s}_n\right)^2
\end{equation}
\end{linenomath}
where $\bar{s}_p = \frac{1}{N_p}\sum_{i=1}^{N_p} s_p^i$ and $\bar{s}_n = \frac{1}{N_n}\sum_{j=1}^{N_n} s_n^j$ represent the mean positive and negative similarity scores, respectively, and $\varepsilon_{\text{pos}} = \varepsilon_{\text{neg}} = 0.01$ are small epsilon values that define tolerance margins. The total variance loss is

\begin{linenomath}
\begin{equation}\label{eq:var_loss}
\mathcal{L}_{\text{var}} = \frac{1}{N_p}\sum_{i=1}^{N_p} \mathcal{L}_{\text{var}}^{\text{pos}} + \frac{1}{N_n}\sum_{j=1}^{N_n} \mathcal{L}_{\text{var}}^{\text{neg}}
\end{equation}
\end{linenomath}

This formulation penalizes positive pairs with similarity below $(1 - \varepsilon_{\text{pos}})\bar{s}_p$ and negative pairs with similarity above $(1 + \varepsilon_{\text{neg}})\bar{s}_n$, thereby reducing intra-class variance and increasing inter-class separation.

\textbf{Combined Loss Function.} The overall training objective combines both loss components with respective weight coefficients:

\begin{linenomath}
\begin{equation}\label{eq:total_loss}
\mathcal{L}_{\text{total}} = \lambda_1 \mathcal{L}_{\text{triplet}} + \lambda_2 \mathcal{L}_{\text{var}}
\end{equation}
\end{linenomath}
where $\lambda_1 = 1.0$ and $\lambda_2 = 0.5$, indicating that identity separation receives higher priority than intra-identity consistency. Together, these components optimize the feature space to produce compact clusters for each animal while maintaining large separation between different identities.

\subsection{{t-SNE Computation}}

To generate the t-distributed Stochastic Neighbor Embedding (t-SNE) visualizations~\citep{vandermaaten2008visualizing}, embeddings from each dataset configuration were extracted and filtered to retain only samples belonging to the top 30 most frequent classes; for each class, a maximum of 100 samples were retained to ensure balanced representation. The high-dimensional embeddings were then projected into a two-dimensional space using t-SNE~\citep{vandermaaten2008visualizing} with perplexity set to 30. Each point in the resulting scatter plot corresponds to an individual sample, colored according to its class identity that ensures visual distinction across all 30~classes. Grid lines and consistent axis scaling were applied across subplots to facilitate direct comparison of cluster structure.

\subsection{{Inference and Evaluation}}

\textbf{Inference Configuration.} Inference uses a batch size of 128 and 8 workers for data loading. Image embeddings are extracted from vision and text encoders and stored in \texttt{pickle} format for efficient retrieval during evaluation.

\textbf{Evaluation Protocol.} Positive pairs (same animal) are generated with constraints: no single image appears more than 5 times across all pairs, and each identity has a maximum of 15 pairs. Negative pairs (different animals) are generated with the same constraints while accounting for image usage in positive pairs. This controlled generation ensures consistent evaluation across all methods. These constraints are necessary to ensure that pair-based verification metrics (Equal Error Rate, Receiver Operating Characteristic Area Under the Curve) accurately reflect model performance rather than artifacts of data imbalance or repeated imagery.

\textbf{Evaluation Metrics.} We report three metrics that assess different aspects of identification performance.

{{Top-k Accuracy} 
} measures the percentage of queries where the correct identity appears in the top $k$ predictions:

\begin{equation}
\text{Top-}k = \frac{\text{Number of correct predictions in top-}k}{\text{Total predictions}}
\end{equation}

We report Top-1, Top-5, and Top-10 accuracy.

\textit{ROC AUC (Receiver Operating Characteristic Area Under the Curve)} measures overall separability of same-animal and different-animal pairs across all decision thresholds:

\begin{equation}
\text{ROC AUC} = \int_0^1 \text{TPR}(\text{FPR}^{-1}(\tau)) \, d\tau
\end{equation}
where TPR is the true positive rate, and FPR is the false positive rate.

\textit{EER (Equal Error Rate)} represents the threshold where the false positive rate equals the false negative rate:

\begin{equation}
\text{EER} = \min_\theta |\text{FPR}(\theta) - \text{FNR}(\theta)|
\end{equation}
where FPR is false positive rate and FNR is false negative rate.

Lower EER indicates better decision boundary calibration. ROC AUC and EER together provide both discrimination and calibration perspectives on model performance.

\textit{The McNemar test} is used to compare two models evaluated on the same test set by checking whether their proportions of correct predictions differ significantly. For each sample, the outcome of model A (correct/incorrect) and model B (correct/incorrect) forms a $2 \times 2$ contingency table with counts a (both correct), b (A correct, B incorrect), c (A incorrect, B correct), and d (both incorrect). The test focuses on the discordant pairs b and c; under the null hypothesis that both models have the same accuracy, these two counts should \mbox{be similar.}

For b + c, the McNemar statistic is
\begin{equation}
    \chi^2 = \frac{(b-c)^2}{b+c}
\end{equation}
which approximately follows a chi-squared distribution with 1 degree of freedom, and the corresponding \emph{p}-value is obtained from this distribution.

\subsection{Hardware}

All experiments were executed on a single {NVIDIA A100 GPU} (NVIDIA Corporation, Santa Clara, CA, USA) 
 with 40 GB of VRAM and an AMD EPYC 7742 64-core CPU  (Advanced Micro Devices, Inc., Santa Clara, CA, USA)..

\section{Results}

\subsection{{Ablation Studies on Different Data Configurations}}

Table~\ref{tab:baseline_results_overall} presents evaluation metrics for various data setups using the \texttt{CLIP-ViT-Base} model~\citep{radford2021learning}, with results computed across the entire test set. For each configuration, the table lists ROC AUC, EER, Top-1, Top-5, and Top-10 accuracy. \textbf{Bold} font indicates the best result in each column, while \underline{{underline} 
} indicates the second-best value. The listed configurations include different combinations of training data and data augmentation strategies.

\begin{table}[H]
\caption{Performance comparison for ablations across different data setups using the \texttt{CLIP-ViT-Base} model~\citep{radford2021learning}; all metrics are computed on the entire test set.\label{tab:baseline_results_overall}}
  \begin{adjustwidth}{-\extralength}{-0cm}
    \begin{tabular*}{\fulllength}{@{\extracolsep{\fill}} l c c c c c}
      \toprule
      \textbf{Configuration} & \textbf{ROC AUC} & \textbf{EER} & \textbf{Top-1} & \textbf{Top-5} & \textbf{Top-10} \\
      \midrule
      Train data & 0.9557 & 0.1042 & 0.6392 & 0.8088 & 0.8535 \\
      Train data + YOLO Preprocessing & {0.9626} & {0.0978} & 0.6282 & 0.7826 & 0.8256 \\
      Train data + PetFace(train split) & 0.9598 & 0.0982 & {0.6460} & {0.8093} & {0.8504} \\
      Train data + Telegram & 0.9716 & 0.0818 & \textbf{0.6791} & \textbf{0.8329} & \textbf{0.8713} \\
      Train data + Pet911 + Telegram & \underline{0.9729} & \underline{0.0772} & \underline{0.6527} & \underline{0.8139} & \underline{0.8568} \\
      Train data + PetFace (full) + Pet911 + Telegram & \textbf{0.9752} & \textbf{0.0729} & 0.6511 & 0.8122 & 0.8555 \\
      \bottomrule
    \end{tabular*}
  \end{adjustwidth}
\end{table}

\begin{figure}[H]
\begin{adjustwidth}{-\extralength}{0cm}
\centering
\includegraphics[width=18cm]{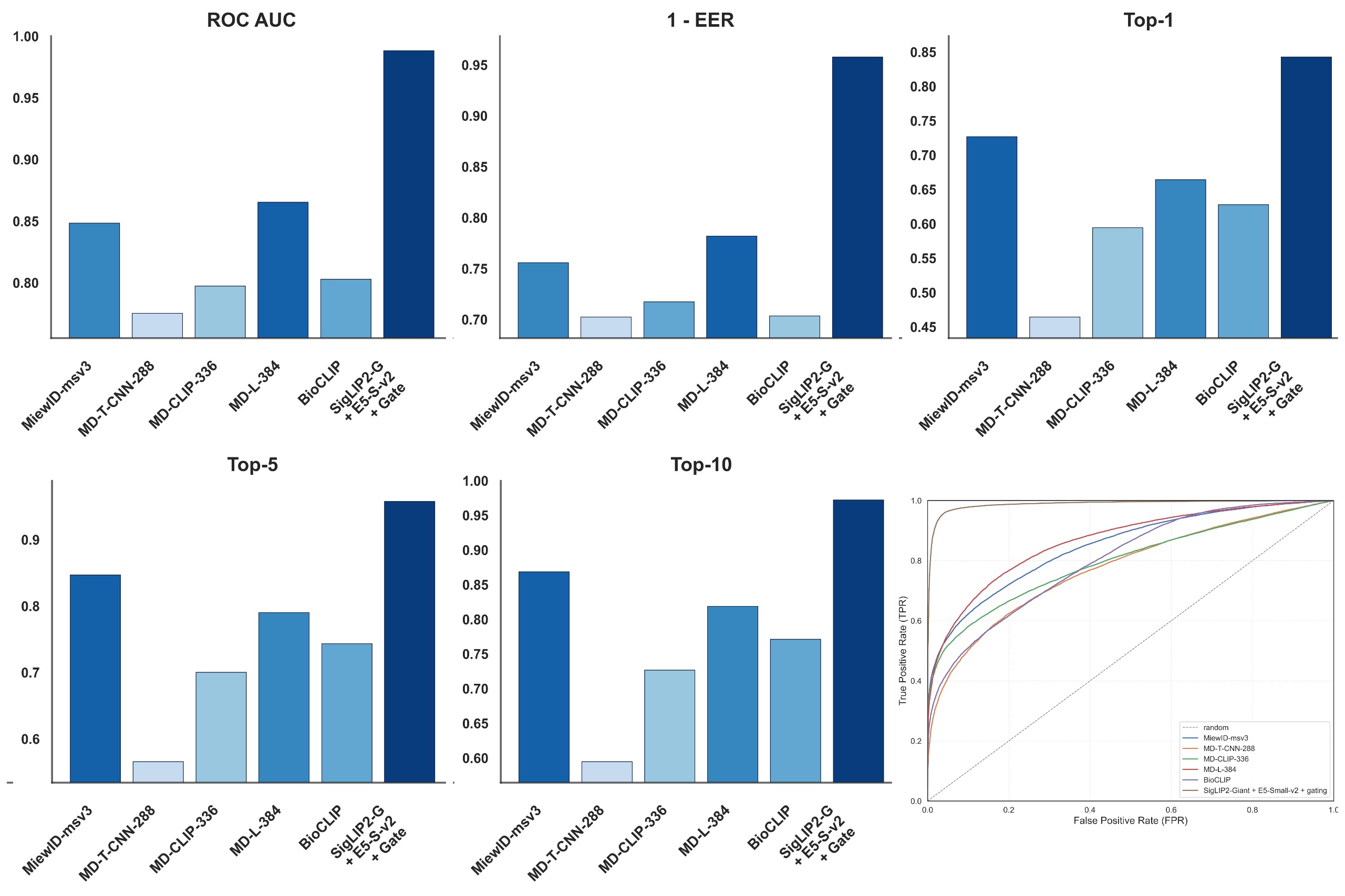}
\end{adjustwidth}
\caption{{{Pet} 
 Identification Performance Across Training Data Configurations: ROC AUC, 1 - EER, and Top-k Metrics.}\label{fig:metrics_data}}
\end{figure}

Figure~\ref{fig:metrics_data} summarizes the quantitative performance of the proposed pet identification models under different training data configurations using a unified visual representation of key metrics. The bar plots report ROC AUC, 1 - EER, and Top‑k identification accuracy for each configuration, while the ROC curves in the bottom‑right panel jointly depict the trade‑off between true positive rate and false positive rate for all evaluated models, including a random baseline. Together, these visualizations consolidate the comparison of alternative training regimes within a single figure and complement the tabulated results by providing an intuitive view of how detection preprocessing and additional data sources affect verification and retrieval behaviour over the entire operating range.

Table~\ref{tab:chi_data} presents the pairwise \emph{p}-values obtained from the McNemar test applied to evaluate the statistical significance of performance differences between models trained under various data configuration settings. Each configuration corresponds to a distinct training dataset composition, including the baseline training data and its combinations with additional sources such as YOLO-based preprocessing, PetFace (training split and full set), Telegram, and Pet911 datasets. The resulting \emph{p}-values provide a comparative measure of whether the observed performance variations between configurations are \mbox{statistically significant.}

Table~\ref{tab:baseline_results_cats} displays performance metrics for various data configurations evaluated on the Cat Individual Images dataset~\citep{timost1234_cat_individuals_2019} using the \texttt{CLIP-ViT-Base} model~\citep{radford2021learning}. Each configuration is assessed using ROC AUC, EER, Top-1, Top-5, and Top-10 accuracy metrics. Values shown in \textbf{bold} represent the highest performance achieved in each respective column, while \underline{underlined} values indicate the second-highest performance.

Table~\ref{tab:baseline_results_dogs} presents performance metrics for several data configurations evaluated using the \texttt{CLIP-ViT-Base} model~\citep{radford2021learning} on the DogFaceNet dataset~\citep{mougeot2019deeplearningapproachdogface}. Configurations are listed alongside ROC AUC, EER, Top-1, Top-5, and Top-10 accuracy values. The best result for each metric column is shown in \textbf{bold}, while the second-best value for each metric is~\underline{underlined}.

\vspace{-9pt}

\begin{table}[H]
\caption{{Pairwise} 
 McNemar test $p$-values for comparing training-data configurations. ↑ indicates fewer errors for the row encoder; {↓ indicates fewer errors for the column encoder.} 
\label{tab:chi_data}}
 \small
  \begin{adjustwidth}{-\extralength}{0cm}
  \begin{tabularx}{\fulllength}{@{}>{\raggedright\arraybackslash}m{5cm}@{}>{\raggedright\arraybackslash}m{2.2cm}@{}>{\raggedright\arraybackslash}m{3.1cm}@{}>{\raggedright\arraybackslash}m{3.2cm}@{}>{\raggedright\arraybackslash}m{2.2cm}@{}>{\raggedright\arraybackslash}m{2.8cm}}
      \toprule
      \textbf{Configuration} &  \textbf{Train Data} &  \textbf{Train Data + YOLO Preprocessing} &  \textbf{Train Data + PetFace (Train Split)} &  \textbf{Train Data + Telegram} &  \textbf{Train Data + Pet911~+ Telegram} \\
      \midrule
      Train data + YOLO Preprocessing & 0.0 (↑) & - & - & - & - \\
      Train data + PetFace (train split) & 0.000012 (↑) & 0.593305 (↑) & - & -& -\\
      Train data + Telegram & 0.0 (↑) & 0.0 (↑) & 0.0 (↑) & - & - \\
      Train data + Pet911 + Telegram &  0.0 (↑) &  0.0 (↑) & 0.0 (↑) & 0.000152 (↑) & -  \\
      Train data + PetFace (full) + \mbox{Pet911~+ Telegram} & 0.0 (↑) & 0.0 (↑) & 0.0 (↑) &  0.0 (↑) &  0.001651 (↑)\\
      \bottomrule
    \end{tabularx}
  \end{adjustwidth}
\end{table}

\vspace{-9pt}

\begin{table}[H]
\caption{Performance comparison for ablations across different data setups using the \texttt{CLIP-ViT-Base} model~\citep{radford2021learning}; all metrics are computed on the  Cat Individual Images~\cite{timost1234_cat_individuals_2019}.\label{tab:baseline_results_cats}}
  \begin{adjustwidth}{-\extralength}{0cm}
    \begin{tabular*}{\fulllength}{@{\extracolsep{\fill}} l c c c c c}
      \toprule
      \textbf{Configuration} & \textbf{ROC AUC}  & \textbf{EER}  & \textbf{Top-1}  & \textbf{Top-5}  & \textbf{Top-10} \\
      \midrule
      Train data & 0.9753 & 0.0761 & 0.8082 & 0.9475 & 0.9655 \\
      Train data + YOLO Preprocessing & {0.9783} & {0.0708} & {0.8411} & 0.9601 & 0.9721 \\
      Train data + PetFace(train split) & 0.9772 & 0.0713 & \underline{0.8491} & \underline{0.9677} & \underline{0.9790} \\
      Train data + Telegram & \textbf{0.9857} & \textbf{0.0528} & \textbf{0.8575} & \textbf{0.9737} & \textbf{0.9832} \\
      Train data + Pet911 + Telegram & \underline{0.9822} & \underline{0.0593} & 0.8296 & 0.9561 & 0.9712 \\
      Train data + PetFace (full) + Pet911 + Telegram & 0.9821 & 0.0604 & 0.8359 & 0.9579 & 0.9711 \\
      \bottomrule
    \end{tabular*}
  \end{adjustwidth}
\end{table}

\begin{table}[H]
\caption{Performance comparison for ablations across different data setups using the \texttt{CLIP-ViT-Base} model~\citep{radford2021learning}; all metrics are computed on the DogFaceNet dataset~\cite{mougeot2019deeplearningapproachdogface}.\label{tab:baseline_results_dogs}}
  \begin{adjustwidth}{-\extralength}{0cm}
    \begin{tabular*}{\fulllength}{@{\extracolsep{\fill}} l c c c c c}
      \toprule
      \textbf{Configuration}  & \textbf{ROC AUC}  & \textbf{EER}  & \textbf{Top-1}  & \textbf{Top-5}  & \textbf{Top-10} \\
      \midrule
      Train data & 0.9653 & 0.0920 & {0.4408} & {0.6477} & \underline{0.7241} \\
      Train data + YOLO Preprocessing & 0.9494 & 0.1207 & 0.3802 & 0.5772 & 0.6575 \\
      Train data + PetFace(train split) & \underline{0.9735} & \underline{0.0774} & 0.4098 & 0.6264 & 0.7029 \\
      Train data + Telegram & 0.9715 & {0.0866} & \textbf{0.4705} & \textbf{0.6688} & \textbf{0.7415} \\
      Train data + Pet911 + Telegram & 0.9727 & 0.0778 & \underline{0.4460} & \underline{0.6479} & 0.7239 \\
      Train data + PetFace (full) + Pet911 + Telegram & \textbf{0.9739} & \textbf{0.0772} & 0.4350 & 0.6417 & 0.7204 \\
      \bottomrule
    \end{tabular*}
  \end{adjustwidth}
\end{table}

\begin{figure}[H]
\begin{adjustwidth}{-\extralength}{0cm}
\centering
\includegraphics[width=15cm]{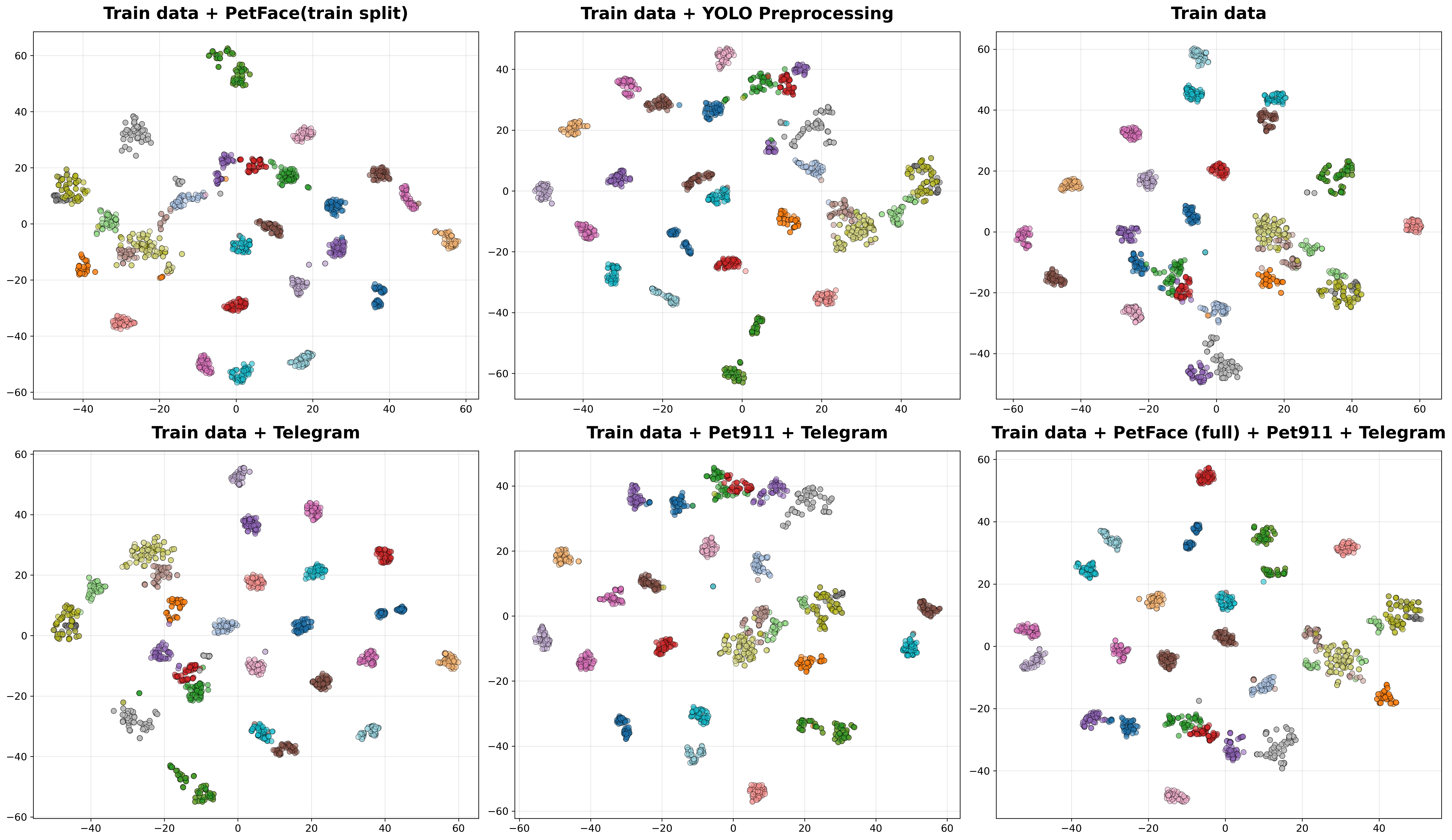}
\end{adjustwidth}
\caption{{{t-SNE} 
 visualization of training data embeddings under different data configurations.}\label{fig:tsne_data}}
\end{figure}

Figure~\ref{fig:tsne_data} presents a comparative t-SNE visualization of data embeddings across six distinct data configurations, each displayed in a separate subplot. In each subplot, colored clusters represent embedded samples projected into a 2D space, with color coding distinguishing different classes or subsets. The spatial distribution and separation of clusters vary across subplots, reflecting how the inclusion of additional datasets or preprocessing steps alters the geometric structure of the feature space. All plots share identical axis ranges for direct visual comparison, and grid lines are included to aid in positional reference.

\subsection{{Vision Encoder Performance Comparison}}

Table~\ref{tab:encoder_results_overall} reports overall performance metrics for six vision encoders evaluated on an identification task, including ROC AUC, EER, and Top-1, Top-5, and Top-10 accuracy. The best-performing value in each column is highlighted in \textbf{bold}, while the second-best value is \underline{underlined} to facilitate rapid comparison across models. All metrics are presented as numerical scores with higher values indicating better performance for ROC AUC and top-k accuracies, and lower values indicating better performance for EER.

\begin{figure}[H]
\begin{adjustwidth}{-\extralength}{0cm}
\centering
\includegraphics[width=18cm]{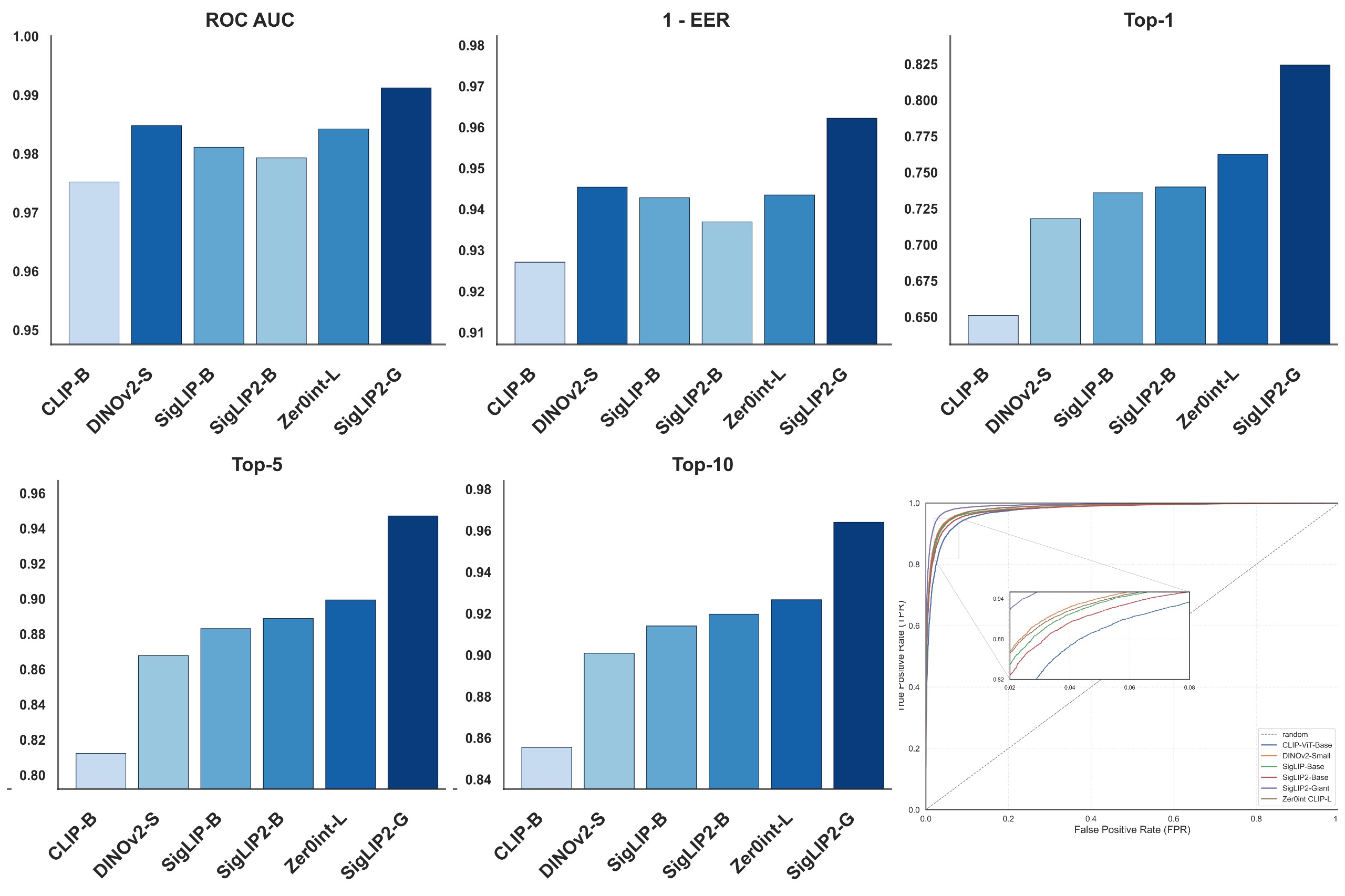}
\end{adjustwidth}
\caption{{{Pet} 
 Identification Performance Across Vision Encoder Configurations: ROC AUC, 1 - EER, and Top-k Metrics.}\label{fig:metrics_vision}}
\end{figure}

Figure~\ref{fig:metrics_vision} provides a compact overview of the evaluation outcomes for all considered vision encoders on the pet identification task. The grouped bar charts report ROC AUC, 1 - EER, and Top-k identification accuracy for each backbone, while the ROC panel on the bottom‑right jointly presents the corresponding verification curves together with a random baseline. By organizing these complementary metrics in a single layout, the figure visually contrasts the behaviour of different foundation models across operating points and ranking depths, complementing the tabular results with an immediate comparison of backbone‑dependent performance characteristics.

\begin{table}[H]
\caption{Vision Encoder Performance---Overall Metrics.\label{tab:encoder_results_overall}}
    \begin{tabular*}{\textwidth}{@{\extracolsep{\fill}}l c c c c c}
      \toprule
      \textbf{Configuration}  & \textbf{ROC AUC}  & \textbf{EER}  & \textbf{Top-1}  & \textbf{Top-5}  & \textbf{Top-10} \\
      \midrule
      CLIP-ViT-Base & 0.9752 & 0.0729 & 0.6511 & 0.8122 & 0.8555 \\
      DINOv2-Small & \underline{0.9848} & \underline{0.0546} & 0.7180 & 0.8678 & 0.9009 \\
      SigLIP-Base & 0.9811 & 0.0572 & 0.7359 & 0.8831 & 0.9140 \\
      SigLIP2-Base & 0.9793 & 0.0631 & 0.7400 & 0.8889 & 0.9197 \\
      Zer0int CLIP-L & 0.9842 & 0.0565 & \underline{0.7626} & \underline{0.8994} & \underline{0.9267} \\
      SigLIP2-Giant & \textbf{0.9912} & \textbf{0.0378} & \textbf{0.8243} & \textbf{0.9471} & \textbf{0.9641} \\
      \bottomrule
    \end{tabular*}
\end{table}

\vspace{-9pt}

Table~\ref{tab:chi_vision} reports pairwise \emph{p}-values obtained from the McNemar test used to assess the statistical significance of differences between identification outcomes across various vision encoder configurations.

Table~\ref{tab:encoder_results_cats} presents performance metrics for six vision encoders evaluated on the Cat Individual Images dataset~\citep{timost1234_cat_individuals_2019}, reporting ROC AUC, EER, and Top-1, Top-5, and Top-10 accuracy. In each column, the highest-performing value is emphasized in \textbf{bold}, while the second-highest is marked with an \underline{underline} to enable direct visual ranking across models. All scores are numerical and presented without interpretation, with higher values preferred for ROC AUC and top-k accuracies, and lower values preferred for EER.

\begin{table}[H]
\caption{{Pairwise} 
 McNemar test $p$-values for comparing vision encoder configurations. ↑ indicates fewer errors for the row encoder; ↓ indicates fewer errors for the column encoder.\label{tab:chi_vision}}
 \small
  \begin{tabularx}{\textwidth}{@{}>{\raggedright\arraybackslash}m{2.5cm}@{}>{\raggedright\arraybackslash}m{2.45cm}@{}>{\raggedright\arraybackslash}m{2.45cm}@{}>{\raggedright\arraybackslash}m{2.1cm}@{}>{\raggedright\arraybackslash}m{2.1cm}@{}>{\raggedright\arraybackslash}m{2.2cm}}
      \toprule
      \textbf{Configuration}  & \textbf{CLIP-ViT-Base}  & \textbf{DINOv2-Small}  & \textbf{SigLIP-Base}  & \textbf{SigLIP2-Base}  & \textbf{Zer0int CLIP-L} \\
      \midrule
      DINOv2-Small & 0.0 (↑) & - & - & - & - \\
      SigLIP-Base & 0.0 (↑)   & 0.005463 (↓)  &  - & - & -   \\
      SigLIP2-Base & 0.0 (↑)  & 0.0 (↓)  & 0.0 (↓) & - & -   \\
      Zer0int CLIP-L & 0.0 (↑)   & 0.134735 (↓)  & 0.30085 (↓)  & 0.0 (↓) & -  \\
      SigLIP2-Giant & 0.0 (↑) & 0.0 (↑) & 0.0 (↑) & 0.0 (↑) & 0.0 (↑) \\
      \bottomrule
    \end{tabularx}
\end{table}

\vspace{-9pt}

\begin{table}[H]
\caption{Vision Encoder Performance---Cat Individual Images~\cite{timost1234_cat_individuals_2019} Metrics.\label{tab:encoder_results_cats}}
    \begin{tabular*}{\textwidth}{@{\extracolsep{\fill}}l c c c c c}
      \toprule
      \textbf{Configuration}  & \textbf{ROC AUC}  & \textbf{EER}  & \textbf{Top-1}  & \textbf{Top-5}  & \textbf{Top-10} \\
      \midrule
      CLIP-ViT-Base & 0.9821 & 0.0604 & 0.8359 & 0.9579 & 0.9711 \\
      DINOv2-Small & \underline{0.9904} & {0.0422} & 0.8547 & 0.9660 & 0.9764 \\
      SigLIP-Base & 0.9899 & 0.0390 & 0.8649 & 0.9757 & 0.9842 \\
      SigLIP2-Base & 0.9894 & \underline{0.0388} & 0.8660 & 0.9772 & \underline{0.9863} \\
      Zer0int CLIP-L & 0.9881 & 0.0509 & \underline{0.8768} & \underline{0.9767} & 0.9845 \\
      SigLIP2-Giant & \textbf{0.9940} & \textbf{0.0344} & \textbf{0.8899} & \textbf{0.9868} & \textbf{0.9921} \\
      \bottomrule
    \end{tabular*}
\end{table}

Table~\ref{tab:encoder_results_dogs} reports quantitative results for six vision encoders evaluated on the DogFaceNet dataset~\citep{mougeot2019deeplearningapproachdogface}, presenting ROC AUC, EER, and Top-1, Top-5, and Top-10 accuracy scores. The best-performing value in each column is indicated in \textbf{bold}, while the second-best is \underline{underlined} to support immediate comparative assessment. All metrics are reported as raw numerical values with higher values indicating superior performance for ROC AUC and top-k accuracy, and lower values preferred for EER.

\begin{table}[H]
\caption{Vision Encoder Performance---DogFaceNet~\cite{mougeot2019deeplearningapproachdogface} Metrics.\label{tab:encoder_results_dogs}}
   \begin{tabular*}{\textwidth}{@{\extracolsep{\fill}}l c c c c c}
      \toprule
      \textbf{Configuration}  & \textbf{ROC AUC}  & \textbf{EER}  & \textbf{Top-1}  & \textbf{Top-5}  & \textbf{Top-10} \\
      \midrule
      CLIP-ViT-Base & 0.9739 & 0.0772 & 0.4350 & 0.6417 & 0.7204 \\
      DINOv2-Small & \underline{0.9829} & \underline{0.0571} & 0.5581 & 0.7540 & 0.8139 \\
      SigLIP-Base & 0.9792 & 0.0606 & 0.5848 & 0.7746 & 0.8319 \\
      SigLIP2-Base & 0.9776 & 0.0672 & 0.5925 & 0.7856 & 0.8422 \\
      Zer0int CLIP-L & 0.9814 & 0.0625 & \underline{0.6289} & \underline{0.8092} & \underline{0.8597} \\
      SigLIP2-Giant & \textbf{0.9926} & \textbf{0.0326} & \textbf{0.7475} & \textbf{0.9009} & \textbf{0.9316} \\
      \bottomrule
    \end{tabular*}
\end{table}

\begin{figure}[H]
\begin{adjustwidth}{-\extralength}{0cm}
\centering
\includegraphics[width=15cm]{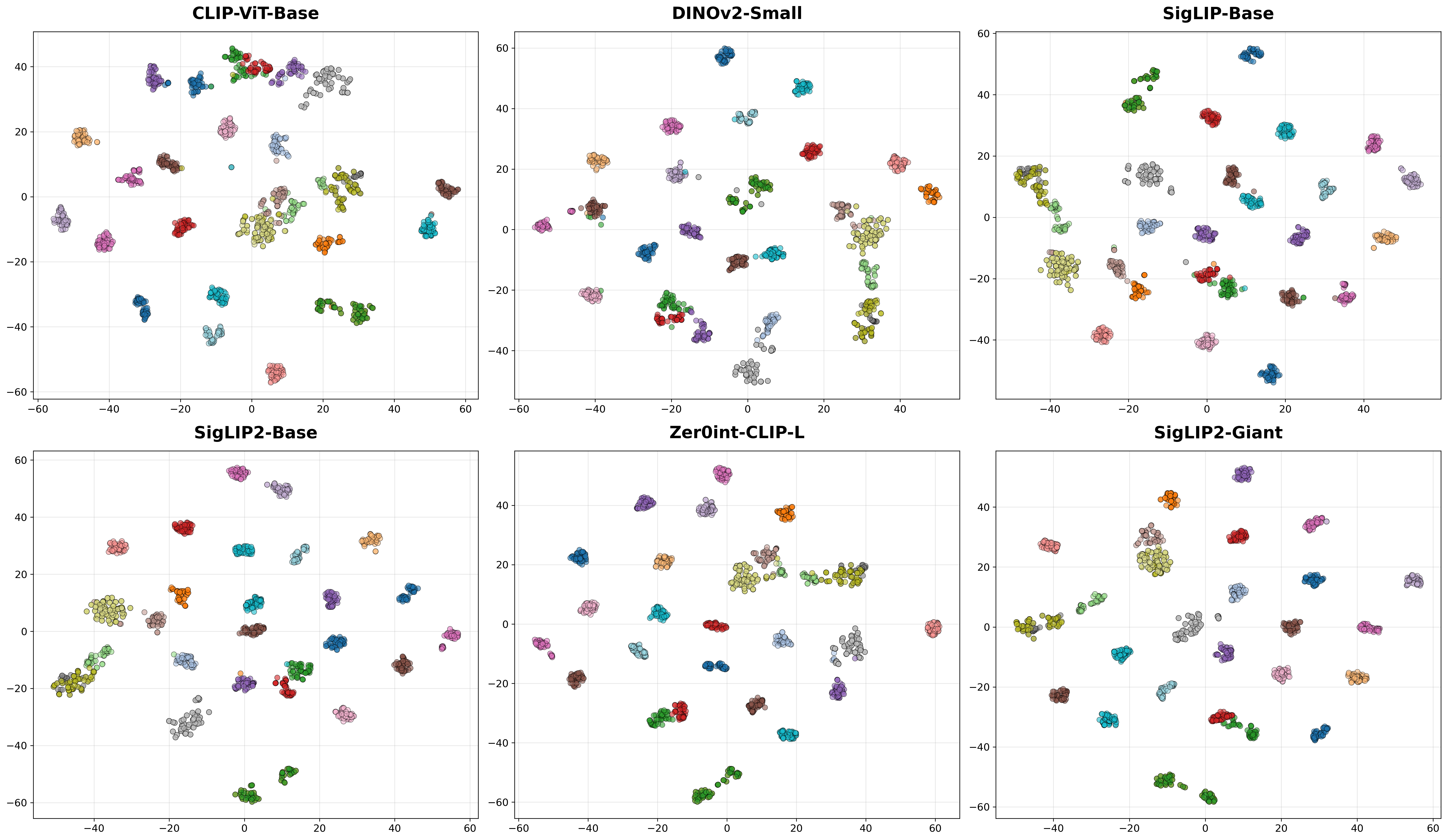}
\end{adjustwidth}
\caption{{{t-SNE} 
 visualization of embeddings under different vision encoder configurations.}\label{fig:tsne_vision}}
\end{figure}

Figure~\ref{fig:tsne_vision} presents t-SNE visualizations of feature embeddings from the six evaluated encoders, computed using samples from the top 30 most frequent classes. The projections reveal varying degrees of structural organization in the latent space; while baseline architectures exhibit noticeable cluster dispersion and inter-class overlap, the larger and specialized models demonstrate notable tighter intra-class grouping and distinct separation. This visual clustering aligns with the quantitative performance, highlighting the superior discriminative capabilities of the advanced architectures.

\subsection{{Text Encoder Performance Comparison}}

Table~\ref{tab:text_results_overall} reports performance metrics for five text encoders evaluated on an overall test set, including ROC AUC, EER, and Top-1, Top-5, and Top-10 accuracy. The best value in each column is indicated in \textbf{bold}, and the second-best value is \underline{underlined} to facilitate direct comparison across models. Higher values are preferable for ROC AUC and top-k accuracy metrics, while lower values are better for EER.

\vspace{-9pt}

\begin{table}[H]
\caption{Text Encoder Performance---Overall Metrics.\label{tab:text_results_overall}}
    \begin{tabular*}{\textwidth}{@{\extracolsep{\fill}}l c c c c c}
      \toprule
      \textbf{Configuration}  & \textbf{ROC AUC}  & \textbf{EER}  & \textbf{Top-1}  & \textbf{Top-5}  & \textbf{Top-10} \\
      \midrule
      BERT & 0.9698 & \textbf{0.0736} & 0.3675 & 0.5717 & 0.6554 \\
      E5-Base & \textbf{0.9703} & \underline{0.0740} & 0.3733 & \underline{0.5748} & {0.6595} \\
      E5-Base-v2 & 0.9697 & 0.0755 & 0.3699 & 0.5743 & 0.6575 \\
      E5-Small & {0.9693} & 0.0764 & \textbf{0.3795} & \textbf{0.5783} & \textbf{0.6615} \\
      E5-Small-v2 & \underline{0.9700} & 0.0757 & \underline{0.3736} & 0.5731 & \underline{0.6597} \\
      \bottomrule
    \end{tabular*}
\end{table}

\begin{figure}[H]
\begin{adjustwidth}{-\extralength}{0cm}
\centering
\includegraphics[width=18cm]{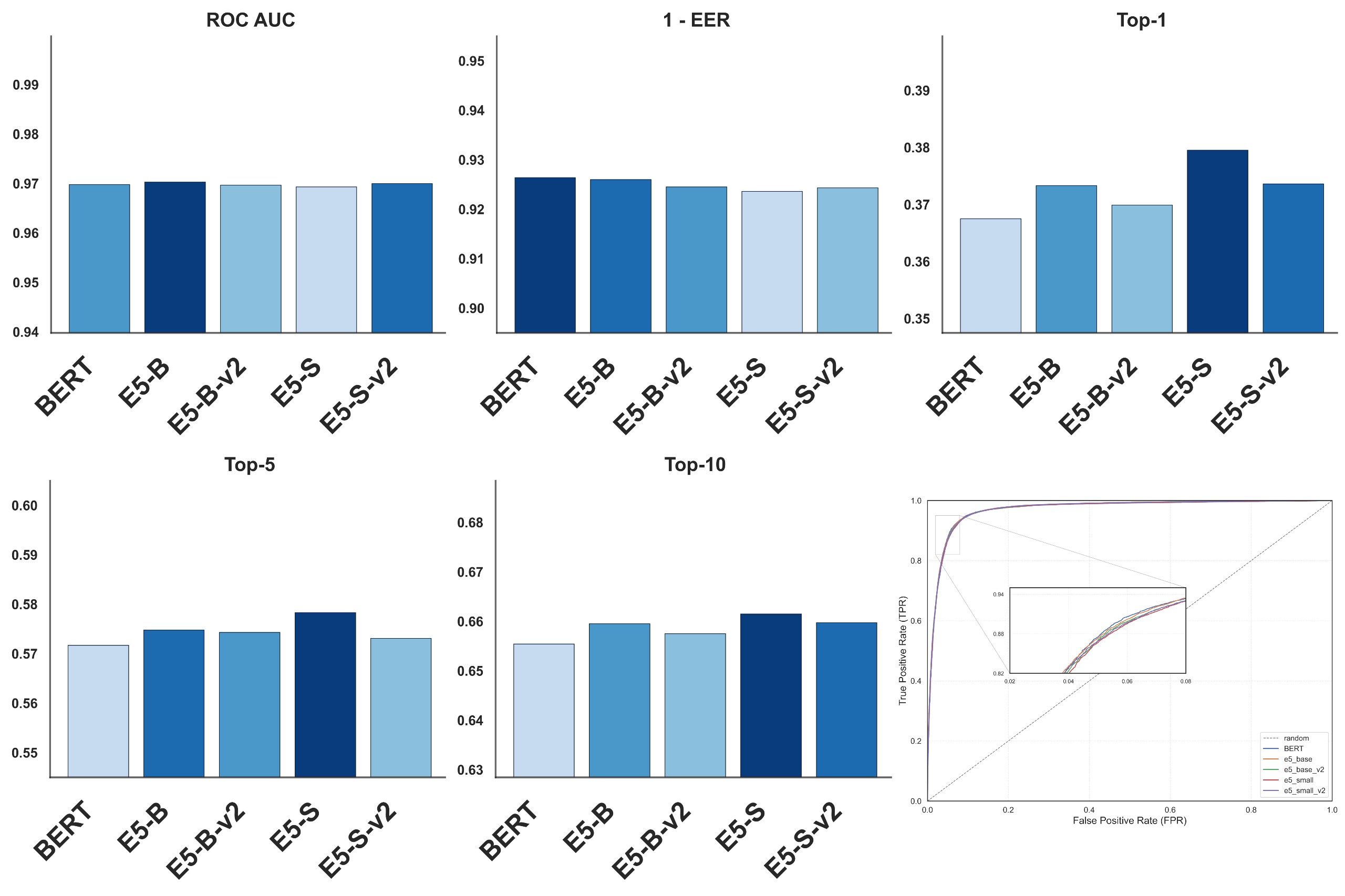}
\end{adjustwidth}
\caption{{{Pet} 
 Identification Performance Across Text Encoder Configurations: ROC AUC, 1 - EER, and Top-k Metrics.}\label{fig:metrics_text}}
\end{figure}

Figure~\ref{fig:metrics_text} illustrates the behaviour of the text encoders on the pet identification benchmark using complementary verification and retrieval metrics. The bar plots summarize ROC AUC, 1 - EER, and Top‑k identification accuracy for BERT and all E5 variants, while the ROC panel compares their corresponding receiver operating characteristic curves with a random baseline over the full false‑positive‑rate range. By presenting these measures side by side, the figure visually contrasts how different text backbones influence matching quality and ranking performance.

Table~\ref{tab:chi_text} presents the pairwise \emph{p}-values obtained from the McNemar test used to assess the statistical significance of differences between various text encoder configurations.  Each cell reports the \emph{p}-value corresponding to the pairwise comparison between two encoders, where lower values indicate stronger evidence of a difference in their performance distributions, while values approaching one suggest similarity.

\begin{table}[H]
\caption{{Pairwise} 
 McNemar test $p$-values for comparing text encoder configurations. ↑ indicates fewer errors for the row encoder; ↓ indicates fewer errors for the column encoder.\label{tab:chi_text}}
    \begin{tabular*}{\textwidth}{@{\extracolsep{\fill}} lllll}
      \toprule
      \textbf{Configuration}  & \textbf{BERT}  & \textbf{ E5-Base }  & \textbf{E5-Base-v2}  & \textbf{E5-Small} \\
      \midrule
      E5-Base & 0.696150 (↓)& - & - & -\\
      E5-Base-v2 & 0.013254 (↓) & 0.029675 (↓) & - & - \\
      E5-Small & 0.044146 (↑) & 0.093004 (↓) & 0.771638 (↑) & - \\
      E5-Small-v2 & 0.016518 (↑) & 0.041364 (↓) & 1.000000 (↓) & 0.731556 (↓)  \\
      \bottomrule
    \end{tabular*}
\end{table}

\vspace{-9pt}

Table~\ref{tab:text_results_cats} reports performance metrics for five text encoders evaluated on the Cat Individual Images dataset~\citep{timost1234_cat_individuals_2019}, including ROC AUC, EER, and Top-1, Top-5, and Top-10 accuracy. The best value in each column is displayed in \textbf{bold}, and the second-best value is \underline{underlined} to support rapid visual comparison across models. Higher values are preferred for ROC AUC and top-k accuracy, while lower values are preferred for EER.

\begin{table}[H]
\caption{Text Encoder Performance---Cat Individual Images~\cite{timost1234_cat_individuals_2019} Metrics.\label{tab:text_results_cats}}
    \begin{tabular*}{\textwidth}{@{\extracolsep{\fill}}l c c c c c}
      \toprule
      \textbf{Configuration}  & \textbf{ROC AUC}  & \textbf{EER}  & \textbf{Top-1}  & \textbf{Top-5}  & \textbf{Top-10} \\
      \midrule
      BERT & \underline{0.9707} & \textbf{0.0726} & 0.5301 & 0.7397 & 0.8060 \\
      E5-Base & 0.9706 & \underline{0.0739} & \underline{0.5396} & {0.7464} & \underline{0.8157} \\
      E5-Base-v2 & \textbf{0.9708} & 0.0751 & 0.5386 & \underline{0.7484} & 0.8131 \\
      E5-Small & 0.9693 & 0.0764 & \textbf{0.5510} & \textbf{0.7504} & \textbf{0.8188} \\
      E5-Small-v2 & 0.9700 & 0.0754 & 0.5373 & 0.7439 & 0.8146 \\
      \bottomrule
    \end{tabular*}
\end{table}

Table~\ref{tab:text_results_dogs} presents performance metrics for five text encoders evaluated on the DogFaceNet dataset~\cite{mougeot2019deeplearningapproachdogface}, including ROC AUC, EER, and Top-1, Top-5, and Top-10 accuracy. The highest-performing value in each column is shown in \textbf{bold}, while the second-highest is \underline{underlined} to enable direct comparison across models. Higher values are preferred for ROC AUC and top-k accuracy, and lower values are preferred for EER.

\begin{figure}[H]
\begin{adjustwidth}{-\extralength}{0cm}
\centering
\includegraphics[width=15cm]{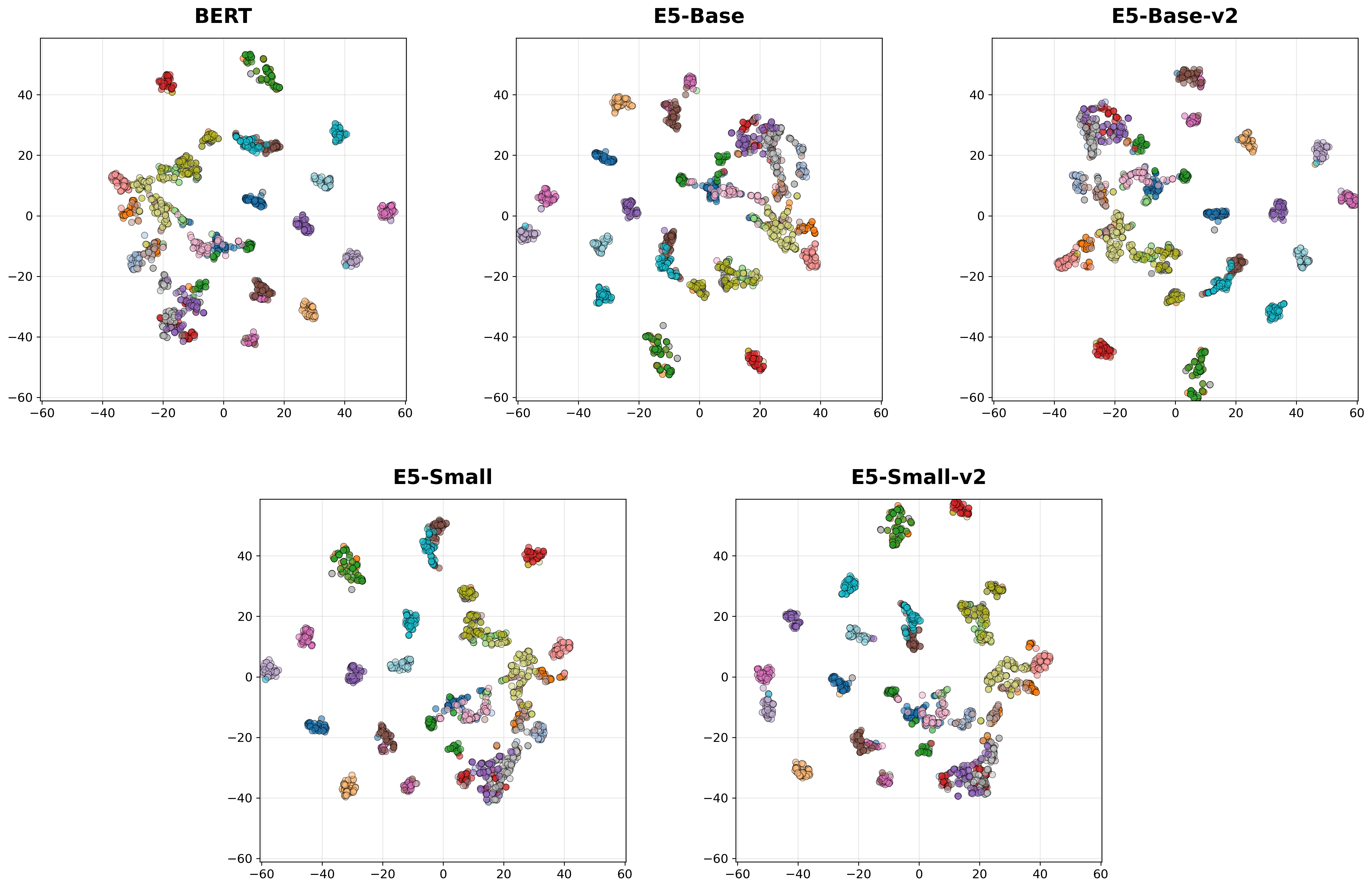}
\end{adjustwidth}
\caption{{{t-SNE} 
 visualization of embeddings under different text encoder configurations.}\label{fig:text_encoder_tsne}}
\end{figure}

Figure~\ref{fig:text_encoder_tsne} shows t-SNE visualizations of text embeddings produced by five different encoder variants. Each subplot presents the distribution of individual clusters corresponding to text descriptions, with distinct colors representing different identities. All visualizations are plotted in a shared coordinate range for direct comparison. The arrangement highlights the separation and clustering patterns achieved by each text encoder across the dataset.

\begin{table}[H]
\caption{Text Encoder Performance---DogFaceNet~\cite{mougeot2019deeplearningapproachdogface} Metrics.\label{tab:text_results_dogs}}
    \begin{tabular*}{\textwidth}{@{\extracolsep{\fill}}l c c c c c}
      \toprule
      \textbf{Configuration}  & \textbf{ROC AUC}  & \textbf{EER}  & \textbf{Top-1}  & \textbf{Top-5}  & \textbf{Top-10} \\
      \midrule
      BERT & 0.9756 & \underline{0.0674} & 0.1776 & \underline{0.3745} & \textbf{0.4791} \\
      E5-Base & \underline{0.9758} & \underline{0.0674} & 0.1789 & 0.3731 & 0.4765 \\
      E5-Base-v2 & \textbf{0.9759} & \textbf{0.0676} & 0.1721 & 0.3713 & 0.4751 \\
      E5-Small & 0.9745 & 0.0685 & \underline{0.1804} & \textbf{0.3761} & 0.4772 \\
      E5-Small-v2 & 0.9753 & {0.0692} & \textbf{0.1821} & 0.3743 & \underline{0.4788} \\
      \bottomrule
    \end{tabular*}
\end{table}

\vspace{-9pt}

\subsection{Comparative Multimodal Performance}

Table \ref{tab:multimodal_results_overall} summarizes the performance of all unimodal and multimodal setups in terms of ROC AUC, EER, and Top-k retrieval accuracy, where \textbf{bold} numbers denote the best result and \underline{underlined} numbers denote the second-best result for each metric.  For fair comparison, the table also reports the corresponding vision-only baselines (\texttt{CLIP-ViT-Base} and \texttt{SigLIP2-Giant}), which serve as reference points for evaluating the added value of the multimodal fusion strategies.

\begin{figure}[H]
\begin{adjustwidth}{-\extralength}{0cm}
\centering
\includegraphics[width=18cm]{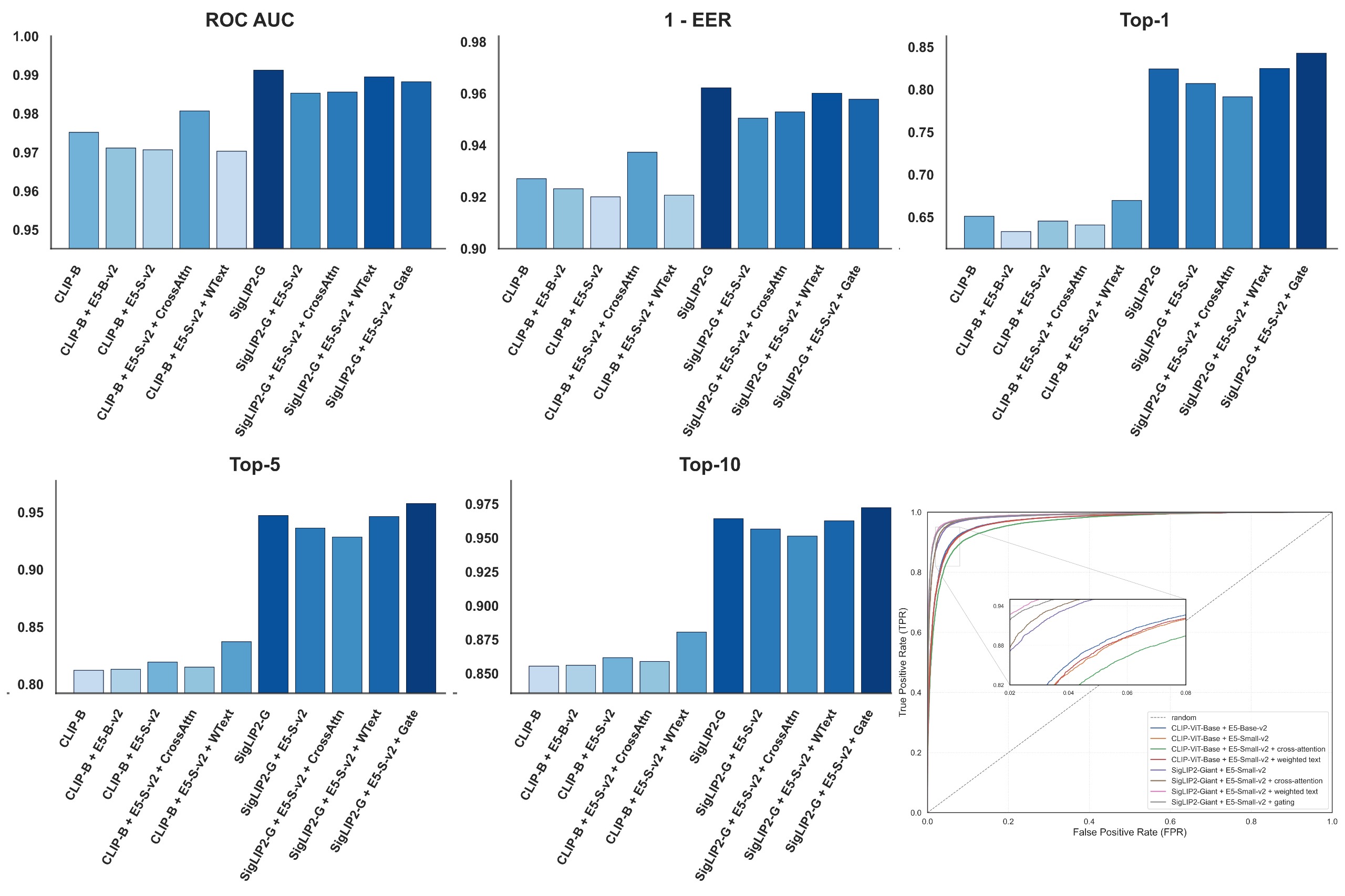}
\end{adjustwidth}
\caption{{{Pet} 
 Identification Performance Across Multimodal Configurations: ROC AUC, 1 - EER, and Top-k Metrics.}\label{fig:metrics_multimodal}}
\end{figure}

Figure~\ref{fig:metrics_multimodal} presents a unified view of the multimodal configurations, combining visual and textual backbones with different fusion strategies. The bar charts summarize ROC AUC, 1 - EER, and Top‑k  identification accuracy for all multimodal variants, while the ROC panel contrasts their receiver operating characteristic curves and a random baseline over the full operating range. This layout makes it possible to visually compare how specific fusion mechanisms and weighting schemes affect both verification quality and retrieval performance across the evaluated multimodal models.

\begin{table}[H]
\caption{Multimodal setups Performance---Overall Metrics.\label{tab:multimodal_results_overall}}
  \begin{adjustwidth}{-\extralength}{0cm}
    \begin{tabular*}{\fulllength}{@{\extracolsep{\fill}}l c c c c c}
      \toprule
      \textbf{Configuration}  & \textbf{ROC AUC}  & \textbf{EER}  & \textbf{Top-1}  & \textbf{Top-5}  & \textbf{Top-10} \\
      \midrule
      CLIP-ViT-Base & 0.9752 & 0.0729 & 0.6511 & 0.8122 & 0.8555 \\
      CLIP-ViT-Base + E5-Base-v2 & 0.9711 & 0.0768 & 0.6331 & 0.8130 & 0.8561 \\
      CLIP-ViT-Base + E5-Small-v2 & 0.9707 & 0.0800 & 0.6454 & 0.8193 & 0.8617 \\
      CLIP-ViT-Base + E5-Small-v2 + cross-attention & 0.9807 & 0.0627 & 0.6409 & 0.8149 & 0.8589 \\
      CLIP-ViT-Base + E5-Small-v2 + weighted text & 0.9703 & 0.0793 & 0.6696 & 0.8371 & 0.8806 \\
      SigLIP2-Giant & \textbf{0.9912} & \textbf{0.0378} & \underline{0.8243} & \underline{0.9471} & \underline{0.9641} \\
      SigLIP2-Giant + E5-Small-v2 & 0.9853 & 0.0495 & 0.8071 & 0.9363 & 0.9566 \\
      SigLIP2-Giant + E5-Small-v2 + cross-attention & 0.9856 & 0.0471 & 0.7916 & 0.9284 & 0.9513 \\
      SigLIP2-Giant + E5-Small-v2 + weighted text & \underline{0.9895} & \underline{0.0399} & 0.8250 & 0.9463 & 0.9625 \\
      SigLIP2-Giant + E5-Small-v2 + gating & 0.9882 & 0.0422 & \textbf{0.8428} & \textbf{0.9576} & \textbf{0.9722} \\
      \bottomrule
    \end{tabular*}
  \end{adjustwidth}
\end{table}

\vspace{-9pt}

Table~\ref{tab:chi_multimodal} presents the pairwise \emph{p}-values obtained using the McNemar test to assess statistical differences in identification outcomes among various multimodal model configurations.  The table provides a comprehensive comparison of the significance levels for all model pairs, enabling evaluation of whether their performance differences are \mbox{statistically distinguishable.}

\begin{table}[H]
\caption{{Pairwise} 
 McNemar test $p$-values for comparing mulimodal models configurations. ↑~indicates fewer errors for the row encoder; ↓ for the column encoder.\label{tab:chi_multimodal}}
\scriptsize
  \begin{adjustwidth}{-\extralength}{0cm}
  \begin{tabularx}{\fulllength}{@{}>{\raggedright\arraybackslash}m{2.5cm}@{}>{\raggedright\arraybackslash}m{2.2cm}@{}>{\raggedright\arraybackslash}m{2.2cm}@{}>{\raggedright\arraybackslash}m{2.4cm}@{}>{\raggedright\arraybackslash}m{2.4cm}@{}>{\raggedright\arraybackslash}m{2.3cm}@{}>{\raggedright\arraybackslash}m{2.4cm}@{}>{\raggedright\arraybackslash}m{2.6cm}}
      \toprule
      \textbf{Configuration}  & \textbf{CLIP-ViT-Base + E5-Base-v2}  & \textbf{ CLIP-ViT-Base + E5-Small-v2 }  & \textbf{CLIP-ViT-Base + E5-Small-v2 + Cross-Attention}  & \textbf{CLIP-ViT-Base + E5-Small-v2 + Weighted Text}  & \textbf{SigLIP2-Giant + E5-Small-v2}  & \textbf{SigLIP2-Giant + E5-Small-v2 + Cross-Attention}  & \textbf{SigLIP2-Giant + E5-Small-v2 + Weighted Text}  \\
      \midrule
      CLIP-ViT-Base + E5-Small-v2 &  0.004174 (↓) & - & - & - & - & - & - \\
      CLIP-ViT-Base + E5-Small-v2 + cross-attention & 0.0 (↓) & 0.0 (↓) & - & - & - & - & -  \\
      CLIP-ViT-Base + E5-Small-v2 + weighted text &0.044237 (↓) & 0.524457 (↑)& 0.0 (↑) & - & - & - & - \\
      SigLIP2-Giant + E5-Small-v2 & 0.0 (↑) &  0.0 (↑) & 0.0 (↑) & 0.0 (↑)  & - & - & - \\
      SigLIP2-Giant + E5-Small-v2 + cross-attention & 0.0 (↑) & 0.0 (↑)& 0.0 (↑) & 0.0 (↑) & 0.002496 (↑) & - & -  \\
      SigLIP2-Giant + E5-Small-v2 + weighted text & 0.0 (↑)& 0.0 (↑) & 0.0 (↑) & 0.0 (↑) & 0.0 (↑) & 0.0 (↑) & -  \\
      SigLIP2-Giant + E5-Small-v2 + gating & 0.0 (↑) & 0.0 (↑) & 0.0 (↑) & 0.0 (↑) & 0.0 (↑) & 0.0 (↑) & 0.000258 (↑)  \\
      \bottomrule
    \end{tabularx}
  \end{adjustwidth}
\end{table}

Table \ref{tab:multimodal_results_cats} summarizes identification performance for the Cat Individual Images dataset~\cite{timost1234_cat_individuals_2019} across different multimodal setups, reporting ROC AUC and EER for verification together with Top-k identification accuracy for a set of image encoders and their multimodal extensions. The table includes \texttt{CLIP-ViT-Base} and \texttt{SigLIP2-Giant} evaluated as pure image encoders, as well as variants that combine these image backbones with E5-based text embeddings using concatenation, cross-attention, weighted-text, and gating fusion strategies, enabling a direct comparison between unimodal image and multimodal image–text configurations under a common evaluation protocol. Values in \textbf{bold} denote the best-performing configuration for each metric column, while \underline{underlined} values indicate the second-best performance in that column.

\begin{table}[H]
\caption{Multimodal setups Performance---Cat Individual Images~\cite{timost1234_cat_individuals_2019} Metrics.\label{tab:multimodal_results_cats}}
  \begin{adjustwidth}{-\extralength}{0cm}
    \begin{tabular*}{\fulllength}{@{\extracolsep{\fill}}l c c c c c}
      \toprule
      \textbf{Configuration}  & \textbf{ROC AUC}  & \textbf{EER}  & \textbf{Top-1}  & \textbf{Top-5}  & \textbf{Top-10} \\
      \midrule
      CLIP-ViT-Base & 0.9821 & 0.0604 & 0.8359 & 0.9579 & 0.9711  \\
      CLIP-ViT-Base + E5-Base-v2 & 0.9824 & 0.0617 & 0.8004 & 0.9445 & 0.9634 \\
      CLIP-ViT-Base + E5-Small-v2 & 0.9839 & 0.0596 & 0.8141 & 0.9532 & 0.9694  \\ 
      CLIP-ViT-Base + E5-Small-v2 + cross-attention & 0.9807 & 0.0627 & 0.8062 & 0.9477 & 0.9669 \\ 
      CLIP-ViT-Base + E5-Small-v2 + weighted text & {0.9843} & {0.0587} & 0.8176 & 0.9497 & 0.9693 \\
      SigLIP2-Giant & \textbf{0.9940} & \underline{0.0344} & 0.8899 & {0.9868} & \underline{0.9921} \\
      SigLIP2-Giant + E5-Small-v2 & 0.9931 & 0.0369 & {0.8867} & 0.9834 & 0.9889 \\
      SigLIP2-Giant + E5-Small-v2 + cross-attention & 0.9922 & 0.0375 & 0.8725 & 0.9779 & 0.9858 \\
      SigLIP2-Giant + E5-Small-v2 + weighted text & \underline{0.9939} & \textbf{0.0315} & \textbf{0.8962} & \textbf{0.9881} & \underline{0.9921} \\
      SigLIP2-Giant + E5-Small-v2 + gating & 0.9929 & \underline{0.0344} & \underline{0.8952} & \underline{0.9872} & \textbf{0.9932} \\
      \bottomrule
    \end{tabular*}
  \end{adjustwidth}
\end{table}

Table \ref{tab:multimodal_results_dogs} reports identification performance on the DogFaceNet benchmark~\citep{mougeot2019deeplearningapproachdogface} across a set of image encoders and their multimodal extensions, summarizing ROC AUC and EER for verification together with Top-k identification accuracy for each encoder configuration. The evaluated models include \texttt{CLIP-ViT-Base} and \texttt{SigLIP2-Giant} used as standalone image encoders for direct comparison, as well as variants that combine these image features with E5-based text embeddings via simple fusion, cross-attention, weighted-text, and gating mechanisms. Values typeset in \textbf{bold} denote the best performance achieved in each metric column, while \underline{underlined} values indicate the second-best performance per column.

\begin{table}[H]
\caption{Multimodal setups Performance---DogFaceNet~\cite{mougeot2019deeplearningapproachdogface} Metrics.\label{tab:multimodal_results_dogs}}
  \begin{adjustwidth}{-\extralength}{0cm}
    \begin{tabular*}{\fulllength}{@{\extracolsep{\fill}}l c c c c c}
      \toprule
      \textbf{Configuration}  & \textbf{ROC AUC}  & \textbf{EER}  & \textbf{Top-1}  & \textbf{Top-5}  & \textbf{Top-10} \\
      \midrule
      CLIP-ViT-Base & 0.9739 & 0.0772 & 0.4350 & 0.6417 & 0.7204 \\
      CLIP-ViT-Base + E5-Base-v2 & 0.9739 & 0.0744 & 0.4371 & 0.6589 & 0.7309\\
      CLIP-ViT-Base + E5-Small-v2 & 0.9717 & 0.0817 & 0.4479 & 0.6628 & 0.7358  \\ 
      CLIP-ViT-Base + E5-Small-v2 + cross-attention & 0.9715 & 0.0836 & 0.4469 & 0.6593 & 0.7326  \\ 
      CLIP-ViT-Base + E5-Small-v2 + weighted text & 0.9786 & 0.0676 & 0.4960 & 0.7054 & 0.7768 \\
      SigLIP2-Giant & \underline{0.9926} & {0.0326} & \underline{0.7475} & \underline{0.9009} & \underline{0.9316} \\
      SigLIP2-Giant + E5-Small-v2 & 0.9909 & 0.0368 & 0.7141 & 0.8810 & 0.9190\\
      SigLIP2-Giant + E5-Small-v2 + cross-attention & 0.9898 & 0.0380 & 0.6964 & 0.8706 & 0.9112   \\
      SigLIP2-Giant + E5-Small-v2 + weighted text & \textbf{0.9932} & \textbf{0.0309} & 0.7420 & 0.8975 & 0.9280 \\
      SigLIP2-Giant + E5-Small-v2 + gating & 0.9920 & \underline{0.0314} & \textbf{0.7818} & \textbf{0.9233} & \textbf{0.9482}  \\
      \bottomrule
    \end{tabular*}
  \end{adjustwidth}
\end{table}

\begin{figure}[H]
\begin{adjustwidth}{-\extralength}{0cm}
\centering
\includegraphics[width=15cm]{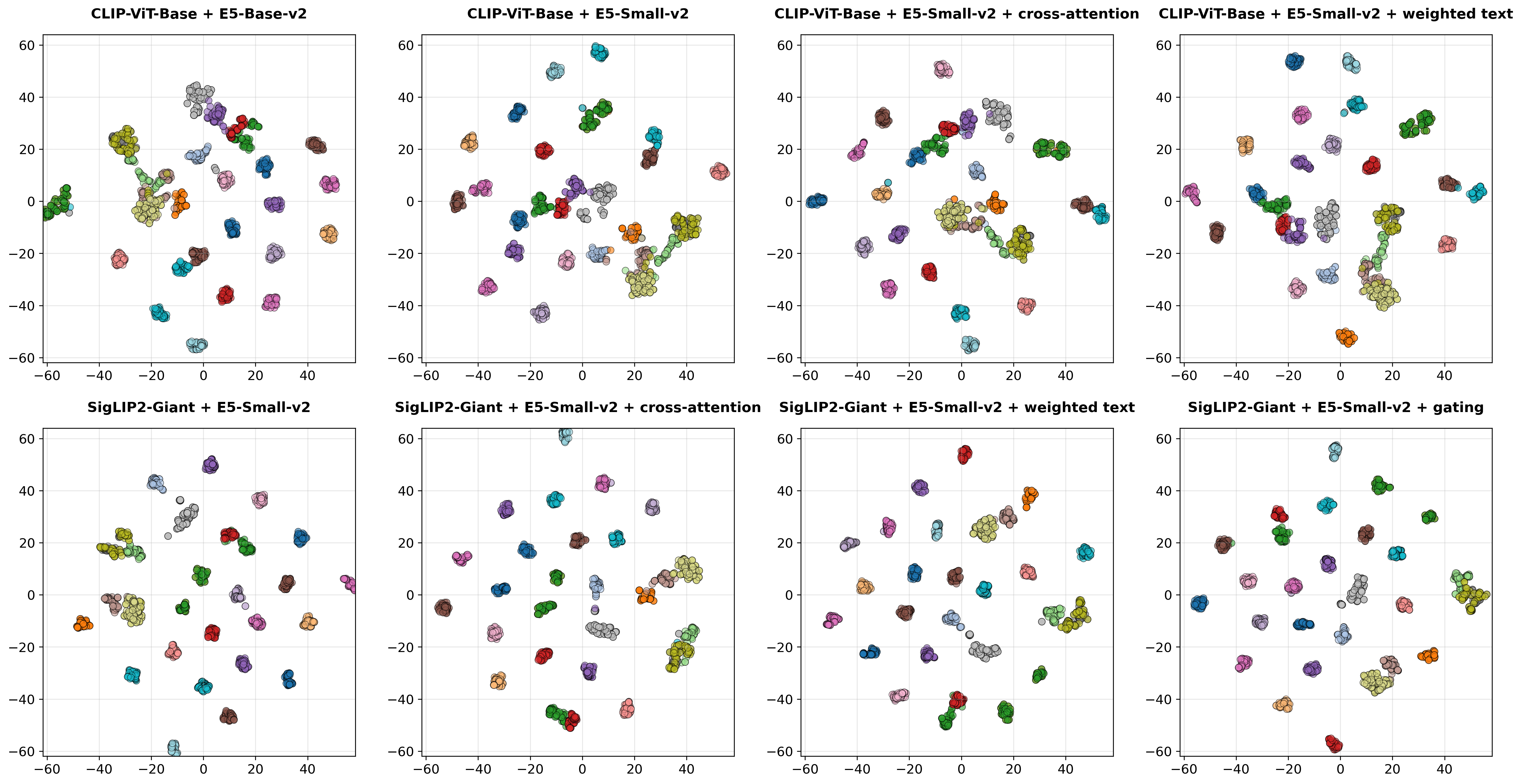}
\end{adjustwidth}
\caption{{{t-SNE} 
 visualization of embeddings under different multimodal setup configurations.}\label{fig:multimodal_encoder_tsne}}
\end{figure}

Figure~\ref{fig:multimodal_encoder_tsne} represents the t-SNE visualization of embeddings generated by the \texttt{CLIP-ViT-Base} (top row) and \texttt{SigLIP2-Giant} (bottom row) backbones across the different evaluated multimodal configurations. Distinct colors correspond to individual identities, illustrating the variations in class separability and cluster compactness achieved by each fusion strategy in the shared embedding space.

\subsection{Comparison of the Proposed Methods with Existing Approaches}

Table \ref{tab:comparison_overall} presents a quantitative comparison of the proposed multimodal approach against several existing methods, summarizing performance metrics such as ROC AUC and EER for verification, as well as Top-k identification accuracy. The proposed configuration (\texttt{SigLIP2-Giant + E5-Small-v2 + gating}) is benchmarked against \texttt{MiewID-msv3}, three variants of the MegaDescriptor family (\texttt{MD-T-CNN-288}, \texttt{MD-CLIP-336}, \texttt{MD-L-384}), and \texttt{BioCLIP}. \textbf{Bold} values indicate the best performance achieved in each column, while \underline{underlined} values denote the second-best results.

\begin{table}[H]
\caption{Comparison of the proposed methods with existing approaches---Overall Metrics.\label{tab:comparison_overall}}
  \begin{adjustwidth}{-\extralength}{0cm}
    \begin{tabular*}{\fulllength}{@{\extracolsep{\fill}}l c c c c c}
      \toprule
      \textbf{Configuration}  & \textbf{ROC AUC}  & \textbf{EER}  & \textbf{Top-1}  & \textbf{Top-5}  & \textbf{Top-10} \\
      \midrule
      MiewID-msv3 & 0.8484 & 0.2442 & \underline{0.7270} & \underline{0.8468} & \underline{0.8685} \\
      MiewID-msv3 & \underline{0.9227} & \underline{0.1508} & \underline{0.5569} & \underline{0.7057} & \underline{0.7447} \\
      MD-T-CNN-288 & 0.7592 & 0.3082 & 0.1599 & 0.2515 & 0.2955\\
      MD-CLIP-336 & 0.7767 & 0.2998 & 0.3410 & 0.4648 & 0.5138 \\
      MD-L-384 & 0.9070 & 0.1673 & 0.4607 & 0.6170 & 0.6695 \\
      BioCLIP & 0.8383 & 0.2465 & 0.3720 & 0.5047 & 0.5539 \\
      SigLIP2-Giant + E5-Small-v2 + gating & \textbf{0.9920} & \textbf{0.0314} & \textbf{0.7818} & \textbf{0.9233} & \textbf{0.9482}  \\
      \bottomrule
    \end{tabular*}
  \end{adjustwidth}
\end{table}

\begin{figure}[H]
\begin{adjustwidth}{-\extralength}{0cm}
\centering
\includegraphics[width=18cm]{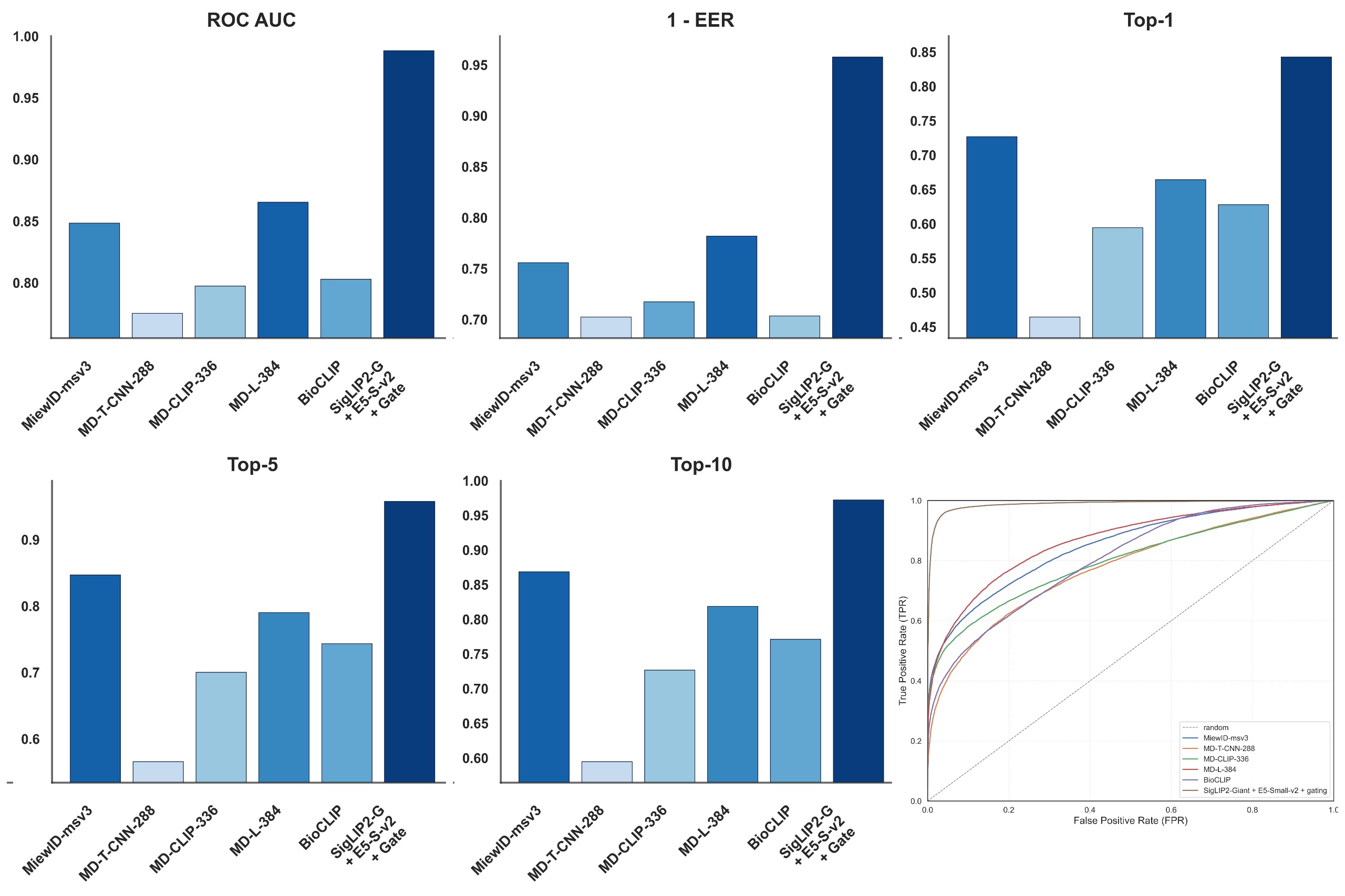}
\end{adjustwidth}
\caption{{{Pet} 
 Identification Performance Across Compared Approaches: ROC AUC, 1 - EER, and Top-k Metrics.}\label{fig:metrics_comparison}}
\end{figure}

Figure~\ref{fig:metrics_comparison} reports the performance of the proposed system in comparison with existing pet re‑identification approaches. The grouped bar charts present ROC AUC, 1 - EER, and Top‑k identification accuracy for prior baselines and for the best configuration of the method introduced in this work, while the ROC panel visualizes their receiver operating characteristic curves together with a random reference. This visualization highlights, in a single view, how the proposed approach compares to established methods in terms of both verification quality and retrieval accuracy across different operating points.

Table~\ref{tab:chi_comparison} reports the pairwise \emph{p}-values from the McNemar test used to evaluate statistical differences in identification outcomes among existing approaches. The results show the significance of performance differences between each pair of models, allowing for assessment of whether the observed improvements are statistically meaningful across the \mbox{compared configurations.}

\begin{table}[H]
\caption{{Pairwise} 
 McNemar test $p$-values for comparing existing approaches. ↑ indicates fewer errors for the row encoder; ↓ indicates fewer errors for the column encoder.\label{tab:chi_comparison}}
  \begin{adjustwidth}{-\extralength}{-0cm}
    \begin{tabular*}{\fulllength}{@{\extracolsep{\fill}} llllll}
      \toprule
      \textbf{Configuration}  & \textbf{MiewID-msv3}  & \textbf{  MD-T-CNN-288 }  & \textbf{MD-CLIP-336}  & \textbf{MD-L-384}  & \textbf{BioCLIP}  \\
      \midrule
      MD-T-CNN-288 & 0.0 (↓) & - & - & - & -\\
      MD-CLIP-336 &  0.0 (↓) & 0.0 (↑) & - & - & -\\
      MD-L-384 & 0.0 (↓)& 0.0 (↓) & 0.0 (↑) &-  & - \\
      BioCLIP & 0.0 (↓)  &  0.0 (↓) &0.0 (↓) & 0.0 (↓) & -\\
      SigLIP2-Giant + E5-Small-v2 + gating &0.0 (↑) & 0.0 (↑) & 0.0 (↑) & 0.0 (↑) & 0.0 (↑) \\
      \bottomrule
    \end{tabular*}
  \end{adjustwidth}
\end{table}

Table \ref{tab:comparison_cats} presents a comparative analysis of the proposed multimodal method against several existing approaches on the Cat Individual Images dataset~\cite{timost1234_cat_individuals_2019}. The evaluations report ROC AUC and EER for verification, along with Top-k identification accuracies for each configuration. \textbf{Bold} values denote the best performance achieved in each metric column, while \underline{underlined} values indicate the second-best results.

\begin{table}[H]
\caption{Comparison of the proposed methods with existing approaches---Cat Individual Images~\cite{timost1234_cat_individuals_2019} Metrics.\label{tab:comparison_cats}}
  \begin{adjustwidth}{-\extralength}{0cm}
    \begin{tabular*}{\fulllength}{@{\extracolsep{\fill}}l c c c c c}
      \toprule
      \textbf{Configuration}  & \textbf{ROC AUC}  & \textbf{EER}  & \textbf{Top-1}  & \textbf{Top-5}  & \textbf{Top-10} \\
      \midrule
      MiewID-msv3 & \underline{0.9383} & \underline{0.1392} & \underline{0.8723} & \underline{0.9695} & \underline{0.9760} \\
      MD-T-CNN-288 & 0.8695 & 0.2111 & 0.7347 & 0.8546 & 0.8743 \\
      MD-CLIP-336 & 0.8953 & 0.1848 & 0.8182 & 0.9158 & 0.9277 \\
      MD-L-384 & 0.9364 & 0.1401 & 0.8421 & 0.9464 & 0.9567 \\
      BioCLIP & 0.8754 & 0.2141 & 0.8473 & 0.9478 & 0.9585 \\
      SigLIP2-Giant + E5-Small-v2 + gating & \textbf{0.9929} & \textbf{0.0344} & \textbf{0.8952} & \textbf{0.9872} & \textbf{0.9932} \\
      \bottomrule
    \end{tabular*}
  \end{adjustwidth}
\end{table}

\textls[-5]{Table \ref{tab:comparison_dogs} details the performance of the proposed multimodal method relative to existing approaches on the DogFaceNet dataset~\cite{mougeot2019deeplearningapproachdogface}, summarizing verification metrics and identification accuracy. \textbf{Bold} formatting highlights the best result in each column, while \underline{underlined} values indicate the second-best performance across the evaluated configurations.}

\begin{table}[H]
\caption{Comparison of the proposed methods with existing approaches---DogFaceNet~\cite{mougeot2019deeplearningapproachdogface} Metrics.\label{tab:comparison_dogs}}
  \begin{adjustwidth}{-\extralength}{0cm}
    \begin{tabular*}{\fulllength}{@{\extracolsep{\fill}}l c c c c c}
      \toprule
      \textbf{Configuration}  & \textbf{ROC AUC}  & \textbf{EER}  & \textbf{Top-1}  & \textbf{Top-5}  & \textbf{Top-10} \\
      \midrule
      MiewID-msv3 & \underline{0.9227} & \underline{0.1508} & \underline{0.5569} & \underline{0.7057} & \underline{0.7447} \\
      MD-T-CNN-288 & 0.7592 & 0.3082 & 0.1599 & 0.2515 & 0.2955\\
      MD-CLIP-336 & 0.7767 & 0.2998 & 0.3410 & 0.4648 & 0.5138 \\
      MD-L-384 & 0.9070 & 0.1673 & 0.4607 & 0.6170 & 0.6695 \\
      BioCLIP & 0.8383 & 0.2465 & 0.3720 & 0.5047 & 0.5539 \\
      SigLIP2-Giant + E5-Small-v2 + gating & \textbf{0.9920} & \textbf{0.0314} & \textbf{0.7818} & \textbf{0.9233} & \textbf{0.9482}  \\
      \bottomrule
    \end{tabular*}
  \end{adjustwidth}
\end{table}

\begin{figure}[H]
\begin{adjustwidth}{-\extralength}{0cm}
\centering
\includegraphics[width=15cm]{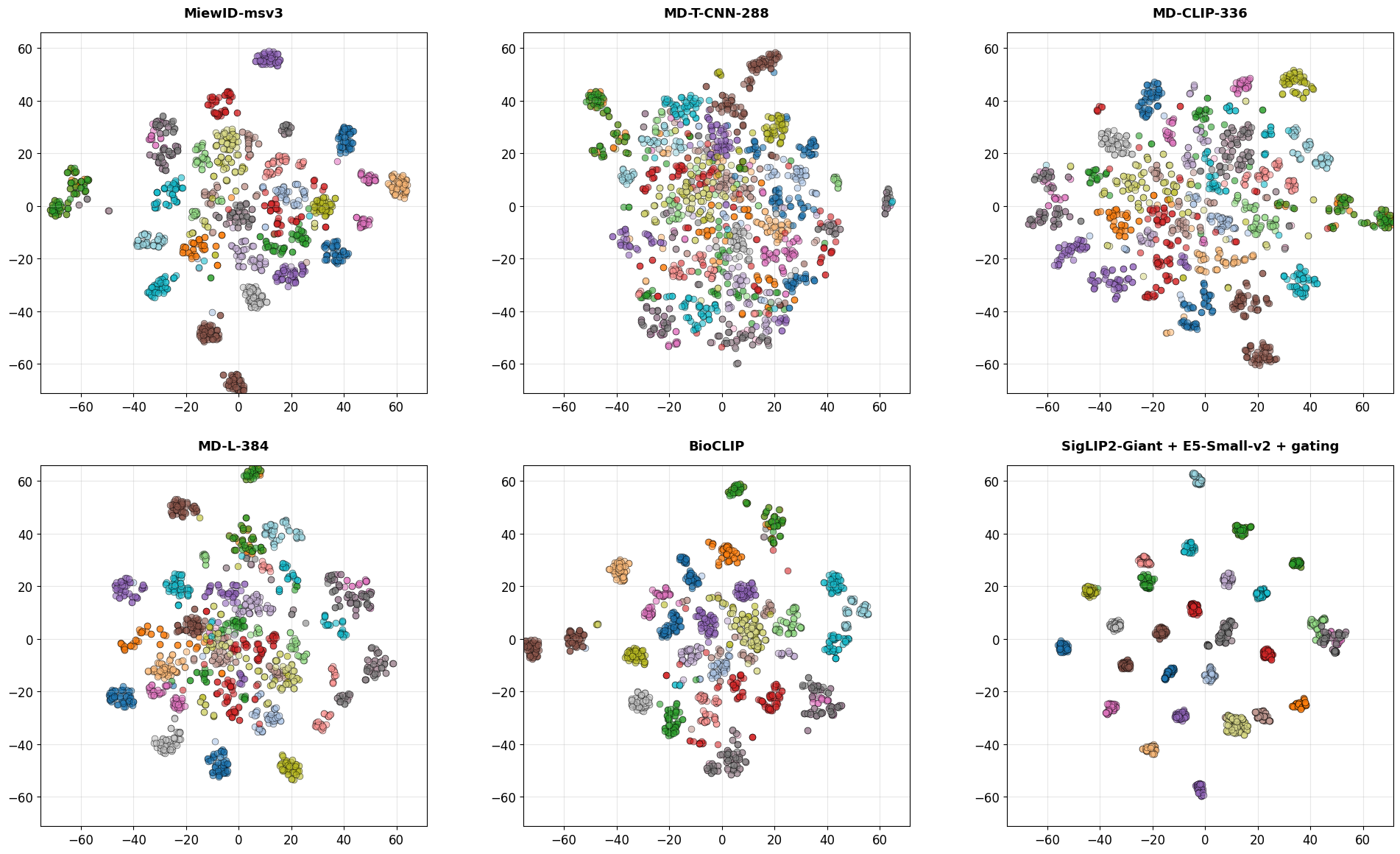}
\end{adjustwidth}
\caption{{{t-SNE} 
 visualization of embeddings under different feature extraction models, comparing the proposed method against existing approaches.}\label{fig:comparison_encoder_tsne}}
\end{figure}

Figure~\ref{fig:comparison_encoder_tsne} depicts t-SNE visualization of feature embeddings generated by the proposed method  (\texttt{SigLIP2-Giant + E5-Small-v2 + gating}) compared to established benchmarks, including \texttt{MiewID-msv3}, variants of the MegaDescriptor family (\texttt{MD-T-CNN-288}, \texttt{MD-CLIP-336}, \texttt{MD-L-384}), and \texttt{BioCLIP}. Distinct colors correspond to individual identities, demonstrating the relative compactness of clusters and the degree of class separability achieved by each approach in the shared embedding space.

\section{Discussion}

\subsection{{Effect of Dataset Configuration Choices}}

Ablation results in Tables~\ref{tab:baseline_results_overall},~\ref{tab:baseline_results_cats},~\ref{tab:baseline_results_dogs} and Figure~\ref{fig:tsne_data} illustrate that the amount and diversity of training data directly impact identification performance. We found that the configuration using the largest, most heterogeneous dataset consistently achieves the highest ROC AUC and lowest EER, though it does not always achieve the highest Top-1 accuracy. This setup outperformed others in threshold-based metrics.

Despite occasional drops in Top-1 and Top-5 scores, incorporating user-generated data sources and expanding data scale resulted in more resilient and generalizable models. Based on these findings, we adopted the most comprehensive data configuration for all subsequent experiments, prioritizing broader robustness over isolated accuracy peaks.

The t-SNE visualization~\citep{vandermaaten2008visualizing} in Figure~\ref{fig:tsne_data} shows that larger, multi-source datasets produce tighter and more separated clusters, supporting our decision to use this configuration. These qualitative results reinforce that expanding dataset diversity improves the reliability of representations, especially when evaluated with robust, threshold-based metrics.

\subsection{{Impact of the Vision Encoder on Task Performance}}

Across all evaluation settings, \texttt{SigLIP2-Giant}~\citep{tschannen2025siglip2multilingualvisionlanguage} establishes a clear performance upper bound, indicating that both model capacity and refined vision–language pre-training translate into stronger animal verification capabilities. On the overall test set, it attains the highest ROC AUC of 0.9912 and the lowest EER of 0.0378, while also surpassing all competing encoders in Top-1, Top-5, and Top-10 accuracy, with Top-1 reaching 0.8243~compared to 0.7626 for the best CLIP-based baseline (\texttt{Zer0int CLIP-L}). This pattern points to more discriminative, well-calibrated decision boundaries that better separate same-animal and different-animal pairs across a wide range of thresholds.

The advantage of \texttt{SigLIP2-Giant}~\citep{tschannen2025siglip2multilingualvisionlanguage} is equally clear in the species-specific benchmarks, where it dominates all metrics on both Cat Individual Images and DogFaceNet. On Cat Individual Images~\citep{timost1234_cat_individuals_2019}, it achieves ROC AUC of 0.9940, EER of 0.0344, outperforming self-supervised \texttt{DINOv2-Small}~\citep{oquab2024dinov2learningrobustvisual} and high-capacity \texttt{Zer0int CLIP-L}, which already operate near the top of the scale. A similar trend holds on DogFaceNet~\citep{mougeot2019deeplearningapproachdogface}, where \texttt{SigLIP2-Giant}~\citep{tschannen2025siglip2multilingualvisionlanguage} reaches ROC AUC of 0.9926, EER of 0.0326, and Top-1 accuracy of 0.7475, confirming that its benefits transfer to both cat and dog identification despite differences in pose and acquisition conditions.

Qualitative t-SNE analysis further supports these quantitative trends, showing that \texttt{SigLIP2-Giant}~\citep{tschannen2025siglip2multilingualvisionlanguage} is the only encoder that produces embeddings with consistently compact, well-separated clusters across all examined settings. In the corresponding subplot, class manifolds occupy distinct, minimally overlapping regions of the 2D space, whereas alternative encoders show noticeable cluster fragmentation and overlap even when their scalar metrics are competitive. This pronounced cluster separability suggests that \texttt{SigLIP2-Giant}~\citep{tschannen2025siglip2multilingualvisionlanguage} learns feature spaces with reduced intra-class variance and increased inter-class margins, which in turn explains its superior metrics and underlines its suitability as a backbone for large-scale individual animal identification.

\subsection{{Impact of the Text Encoder on Task Performance}}\label{discussion:text}

The impact of the text encoder on task performance is distinct from that of the image encoder. Our experiments show that text encoders operating on synthetic descriptions achieve a Top-1 accuracy of 37.95\% on the overall test set, which is lower than the vision-only baseline. However, despite this difference in ranking capability, text encoders maintain robust verification performance, consistently achieving ROC AUC scores around 0.97 across datasets like DogFaceNet~\cite{mougeot2019deeplearningapproachdogface}.  This suggests that while text descriptions may lack the pixel-level granularity required for high-precision ranking, they successfully capture core identity features that remain stable across different instances. This observation is further supported by our t-SNE analysis (Figure~\ref{fig:text_encoder_tsne}), which shows that while text embeddings form broader clusters with only partial separability between identities, which contrasts with the tighter clustering observed in visual features. Notably, this distinct modality provides complementary information, as evidenced by the fact that integrating these text embeddings into a multimodal setup yields the highest overall metrics in our study, Section~\ref{discussion:multimodal}. The varying performance across different text architectures remains negligible, indicating that the utility of the text branch is driven more by the semantic content of the descriptions than by the specific encoder architecture used.

\subsection{Performance Gains from Multimodal Fusion}\label{discussion:multimodal}

Comparison of fusion strategies reveals that combining visual features with text embeddings primarily enhances ranking performance rather than verification metrics. While the standalone \texttt{SigLIP2-Giant} vision encoder retains the highest ROC AUC and EER scores, the multimodal setup notably improves Top-k accuracies. Specifically, the fusion of \texttt{SigLIP2-Giant} with the \texttt{E5-Small-v2} text encoder using a gating mechanism achieves a Top-1 accuracy of 84.28\% on the overall test set, surpassing the 82.43\% accuracy of the vision-only baseline. This trend holds for sub-datasets as well, where the gated fusion model on \texttt{DogFaceNet} reaches a Top-1 accuracy of 78.18\% compared to 74.75\% for the vision baseline. Among the evaluated strategies, the gating mechanism proved most effective, likely because it dynamically assigns importance weights to image and text features. This quantitative improvement is visually corroborated by our t-SNE analysis (Figure~\ref{fig:multimodal_encoder_tsne}), which demonstrates that the multimodal embeddings form more distinct and compact clusters than those of the unimodal baselines, confirming that integrating semantic attributes from text allows the model to better distinguish between similar candidates in ranking tasks.

\subsection{Advantages of the Proposed Multimodal Framework}

Our proposed multimodal framework, specifically the \texttt{SigLIP2-Giant} encoder combined with \texttt{E5-Small-v2} using a gating mechanism, demonstrates improved performance compared to established approaches for animal identification. As shown in Table~\ref{tab:comparison_overall}, our method achieves an overall Top-1 accuracy of 84.28\%, notably exceeding the 72.70\% achieved by the best competing model (\texttt{MiewID-msv3}). This corresponds to an absolute improvement of over 11\% in rank-1 identification capabilities. Furthermore, our approach exhibits robust verification reliability, achieving an EER of 0.0422 compared to 0.2442 for \texttt{MiewID-msv3}, indicating precise calibration of decision boundaries.

The effectiveness of our framework is further corroborated by qualitative analysis using t-SNE visualizations (Figure~\ref{fig:comparison_encoder_tsne}). These plots reveal that our method produces more compact and well-separated clusters for individual identities compared to baseline models. While embeddings from models like \texttt{MD-T-CNN-288} show cluster fragmentation and overlap for visually similar individuals, our fusion strategy successfully segregates these identities, confirming that the combination of high-capacity visual features with semantic text descriptors results in a more discriminative representation space.

\subsection{Limitations and Future Work}

\textbf{Dependence on Synthetic Text.} Our current multimodal framework utilizes text descriptions generated by a vision--language model. This approach ensures consistent, structured captions that align well with visual features during training. However, real-world user queries often consist of unstructured or subjective human descriptions. While the current study establishes the baseline efficacy of using generated text, bridging the domain gap between synthetic captions and natural human language remains an open question. Future work will specifically investigate adaptation strategies to handle noisy, human-written queries in ``lost and found'' scenarios.

\textbf{Computational Complexity and Deployment.} To achieve the highest performance observed in our experiments, we leveraged large-scale backbones, specifically the \texttt{SigLIP2-Giant} vision encoder, which contains approximately 2 billion parameters. Furthermore, the most effective multimodal configurations require textual input; in our pipeline, this text is obtained using a 7-billion-parameter VLM. Consequently, the combined computational footprint of the heavy vision encoder and the requisite text generation process presents challenges for resource-constrained deployment. Future research could explore knowledge distillation to retain these capabilities in lighter architectures suitable for edge devices.

\textbf{Label Noise in Scraped Data.} To facilitate large-scale training, we aggregated over 130,000 images from public sources. Although we applied semi-automated cleaning procedures to filter the data, datasets of this magnitude collected from open platforms may contain label noise, such as duplicate entries or inconsistent labeling. Since metric learning is sensitive to such irregularities, this potential noise could impact the precise estimation of the model's upper performance bound.

\textbf{Data Availability.} The primary contribution of this work is the development of advanced architectural approaches and fusion strategies for animal identification. Due to legal and privacy constraints associated with the source platforms, we are unable to publicly release the aggregated datasets. However, to ensure reproducibility and facilitate further research, we release the full codebase, including trained model weights, training scripts, and testing protocols.

\section{Conclusions}

This study addresses the task of individual animal identification by establishing a large-scale evaluation framework and introducing a multimodal recognition pipeline. Through the construction of a training corpus comprising over 1.9 million photographs across 695,000~identities, including data from Pet911 and Telegram, we observed that dataset diversity is essential for developing models that generalize to unseen animals. Our systematic ablation of vision architectures highlighted the effectiveness of \texttt{SigLIP2-Giant}, which achieved an ROC AUC of 0.9912 and produced compact discriminative feature clusters. Additionally, the assessment of multimodal fusion demonstrated that augmenting visual features with textual descriptions via a gated fusion mechanism improved \mbox{ranking capabilities.}

The proposed configuration resulted in a Top-1 accuracy of 84.28\% on the overall test set, surpassing the MiewID-msv3 baseline by approximately 11\% in identification accuracy, with an Equal Error Rate of 0.0422. These findings suggest that integrating semantic identity priors with visual features helps mitigate ambiguity in retrieval scenarios. Consequently, this methodology provides a validated foundation for automated pet reunification systems, while subsequent work may focus on optimizing these large-scale models for resource-constrained environments.

Future work will focus on evaluating the proposed multimodal identification pipeline in real-world scenarios where the textual modality is provided by users and is therefore noisy, subjective, and potentially incomplete compared to the synthetic descriptions used in our current experiments. Such an evaluation requires deploying the system with verified identities and real outcome logs in order to measure end-to-end effectiveness under operational conditions. We therefore treat these directions as important future work rather than topics that can be properly validated within the present experimental framework.

\vspace{6pt}

\authorcontributions{\textls[-25]{Conceptualization, K.B. (Kirill Bubenchikov), G.M. and A.R.; methodology, V.K. and K.B. (Kirill Borodin); software, V.K. and K.B. (Kirill Borodin); validation, V.K., \mbox{K.B.~(Kirill Borodin),} K.B. (Kirill Bubenchikov), G.M. and A.R.; formal analysis, V.K., \mbox{K.B. (Kirill Borodin),}} \mbox{K.B. (Kirill Bubenchikov),} G.M. and A.R.; investigation, V.K.; resources, V.K. and G.M.; data curation, V.K., K.B. (Kirill Borodin) and G.B.; writing---original draft preparation, K.B. (Kirill Borodin), G.B., K.B. (Kirill Bubenchikov), G.M. and A.R.; writing---review and editing, K.B. (Kirill Borodin), G.B., K.B. (Kirill Bubenchikov), G.M. and A.R.; visualization, V.K., K.B. (Kirill Borodin) and G.B.; supervision, K.B. (Kirill Borodin), K.B. (Kirill Bubenchikov), G.M. and A.R.; project administration, \mbox{K.B. (Kirill Bubenchikov),} G.M. and A.R.; funding acquisition, K.B. (Kirill Bubenchikov), G.M. and A.R. All authors have read and agreed to the published version of the manuscript.}

\funding{This 
 research received no external funding.}

\institutionalreview{Not applicable. 
}

\informedconsent{{Not applicable.} 
}

\dataavailability{{Ours} 
 DINO-v2-small version---\url{https://huggingface.co/AvitoTech/DINO-v2-small-for-animal-identification} (accessed 15 November 2025),\\
 ours Zer0int-CLIP-L version---\url{https://huggingface.co/AvitoTech/Zer0int-CLIP-L-for-animal-identification} (accessed 15 November 2025),\\
 ours SigLIP-Base version---\url{https://huggingface.co/AvitoTech/SigLIP-Base-for-animal-identification} (accessed 15 November 2025),\\
 ours SigLIP2-Base version---\url{https://huggingface.co/AvitoTech/SigLIP2-Base-for-animal-identification} (accessed 15 November 2025),\\
 ours CLIP-ViT-Base version---\url{https://huggingface.co/AvitoTech/CLIP-ViT-base-for-animal-identification} (accessed 15 November 2025),\\
 ours SigLIP2-giant version---\url{https://huggingface.co/AvitoTech/SigLIP2-giant} (accessed 15 November 2025),\\
 SigLIP2-giant + E5-Small-v2 + gating---\url{https://huggingface.co/AvitoTech/SigLIP2-giant-e5small-v2-gating} (accessed 15 November 2025),\\
 BeatifulSoup4---\url{https://pypi.org/project/beautifulsoup4/} (accessed 15 November 2025),\\
 Telethon library---\url{https://docs.telethon.dev/en/stable/} (accessed 15 November 2025),\\
 Targeted telegram channel---\url{https://t.me/HvostatPatrul} (accessed 15 November 2025),\\
 CLIP-ViT-Base---\url{https://huggingface.co/openai/clip-vit-base-patch32} (accessed 15 November 2025),\\
 SigLIP-Base---\url{https://huggingface.co/google/siglip-base-patch16-224} (accessed 15 November 2025),\\
 SigLIP2-Base---\url{https://huggingface.co/google/siglip2-base-patch16-224} (accessed 15 November 2025),\\
 SigLIP2-Giant---\url{https://huggingface.co/google/siglip2-giant-opt-patch16-384} (accessed 15 November 2025),\\
 DINOv2-Small---\url{https://huggingface.co/facebook/dinov2-small} (accessed 15 November 2025),\\
 Zer0int CLIP-L---\url{https://huggingface.co/zer0int/CLIP-GmP-ViT-L-14} (accessed 15 November 2025),\\
 Qwen3-VL---\url{https://huggingface.co/Qwen/Qwen3-VL-8B-Instruct} (accessed 15 November 2025),\\
 E5-Base---\url{https://huggingface.co/intfloat/e5-base} (accessed 15 November 2025),\\
 E5-Small---\url{https://huggingface.co/intfloat/e5-small} (accessed 15 November 2025),\\
 E5-Small-v2---\url{https://huggingface.co/intfloat/e5-small-v2} (accessed 15 November 2025),\\
 E5-Base-v2---\url{https://huggingface.co/intfloat/e5-base-v2} (accessed 15 November 2025),\\
 BERT---\url{http://huggingface.co/google-bert/bert-base-uncased} (accessed 15 November 2025),\\
 MD-T-CNN-288---\url{https://huggingface.co/BVRA/MegaDescriptor-T-CNN-288} (accessed 15 November 2025),\\
 MD-CLIP-336---\url{https://huggingface.co/BVRA/MegaDescriptor-CLIP-336} (accessed 15 November 2025),\\
 MD-L-384---\url{https://huggingface.co/BVRA/MegaDescriptor-L-384} (accessed 15 November 2025),\\
 MiewID-msv3---\url{https://huggingface.co/conservationxlabs/miewid-msv3} (accessed 15 November 2025),\\
 BioCLIP---\url{https://huggingface.co/imageomics/bioclip} (accessed 15 November 2025).}

\acknowledgments{Not applicable.}

\conflictsofinterest{Author Kirill Bubenchikov and Alexander Ryzhkov are employed by the company Avito. The remaining authors declare that the research was conducted in the absence of any commercial or financial relationships that could be construed as a potential conflict of interest.}



\abbreviations{Abbreviations}{
{The} 
 following abbreviations are used in this manuscript:
\\

\noindent 
\begin{tabular}{@{}ll}
ROC AUC & Receiver Operating Characteristic Area Under the Curve\\
EER & Equal Error Rate\\
TPR & True Positive Rate\\
FPR & False Positive Rate \\
FNR & False Negative Rate\\
t-SNE & t-Distributed Stochastic Neighbor Embedding\\
\end{tabular}
}

\begin{adjustwidth}{-\extralength}{0cm}

\reftitle{References}



\PublishersNote{}
\end{adjustwidth}
\end{document}